\documentclass{article}

\usepackage{times}
\usepackage{graphicx} 
\usepackage{multicol}
\usepackage{subcaption}
\usepackage{natbib}

\usepackage{algorithm}
\usepackage{algorithmic}

\usepackage{amssymb}
\usepackage{IEEEtrantools}
\usepackage{mathtools}

\usepackage{caption}

\usepackage[flushleft]{threeparttable}

\usepackage{amsmath}

\usepackage{footmisc}

\usepackage{url}
\usepackage{hyperref}

\usepackage[accepted]{icml2017} 

\usepackage{mathtools}  

\usepackage[colorinlistoftodos,prependcaption,textsize=tiny]{todonotes}

\newcommand{\matr}[1]{\mathbf{#1}}
\RequirePackage{xcolor}
\usepackage{xcolor}
\definecolor{Magenta}{rgb}{1,0.0,0.2}

\usepackage{enumitem}

\newcommand{\specialcell}[2][c]{%
  \begin{tabular}[#1]{@{}c@{}}#2\end{tabular}}

\usepackage{mathtools}
\newcommand{\prob}{\operatorname{p}\probarg}
\DeclarePairedDelimiterX{\probarg}[1]{(}{)}{%
  \ifnum\currentgrouptype=16 \else\begingroup\fi
  \activatebar#1
  \ifnum\currentgrouptype=16 \else\endgroup\fi
}
\newcommand{\innermid}{\nonscript\;\delimsize\vert\nonscript\;}
\newcommand{\activatebar}{%
  \begingroup\lccode`\~=`\|
  \lowercase{\endgroup\let~}\innermid 
  \mathcode`|=\string"8000
}

\icmltitlerunning{Leveraging Node Attributes for Incomplete Relational Data}

\begin{document} 

\twocolumn[
\icmltitle{Leveraging Node Attributes for Incomplete Relational Data}



\icmlsetsymbol{equal}{*}

\begin{icmlauthorlist}
\icmlauthor{He Zhao}{fit}
\icmlauthor{Lan Du}{fit}
\icmlauthor{Wray Buntine}{fit}
\end{icmlauthorlist}

\icmlaffiliation{fit}{Faculty of Information Technology, Monash University, Australia}

\icmlcorrespondingauthor{He Zhao}{he.zhao@monash.edu}

\icmlkeywords{side information, relational learning, Poisson factorisation}

\vskip 0.3in
]



\printAffiliationsAndNotice{}  

\begin{abstract} 

  Relational data are usually highly incomplete in practice, which inspires us
  to leverage side information to improve the
  performance of community detection and link prediction. This paper presents
  a Bayesian probabilistic approach that incorporates various kinds of node
  attributes encoded in binary form in relational models with Poisson likelihood. Our method works flexibly with both directed and
  undirected relational networks. 
  The inference can be done by efficient Gibbs sampling which leverages sparsity of
  both networks and node attributes. Extensive experiments
  show that our models achieve the state-of-the-art link prediction results,
  especially with highly incomplete relational data.
\end{abstract}

\section{Introduction}
Relational learning from network data, particularly with probabilistic
methods, has gained a wide range of applications
such as social network analysis~\cite{xiang2010modeling}, recommender systems~\cite{gopalan2014bayesian}, knowledge
graph completion~\cite{hu2016topic}, and bioinformatics~\cite{huopaniemi2010multivariate}. Generally speaking, the goal
of relational learning is to discover and analyse latent clusters of entities
(i.e., community detection), and predict missing links (i.e., link
prediction). 

The standard approach for modelling relational data is latent factor analysis
via matrix factorisation and its variations. Among the existing approaches, 
Non-negative Matrix Factorisation (NMF) and the Stochastic Block Model (SBM)
are prominent foundational methods. NMF
is usually used to model relationships between two sets of entities such as
users and movies in collaborative filtering~\cite{mnih2008probabilistic}. While developed independently, SBM~\cite{wang1987stochastic,nowicki2001estimation}
can be viewed as an extension
of NMF that introduces a
block matrix to capture the interactions between latent factors. There
have been many Bayesian extensions of these two methods, relaxing the
assumptions and/or introducing extra components, such as the Infinite
Relational Model (IRM)~\cite{kemp2006learning}, the mixture membership
stochastic block model (MMSB)~\cite{airoldi2008mixed}, and the 
non-parametric latent feature models
(NLFM)~\cite{miller2009nonparametric}.
Poisson Factorisation
(PF)~\cite{dunson2005bayesian,zhou2012beta},
is a popular version of NMF which models count data
with convenient statistical properties~\cite{gopalan2014bayesian,Gopalan2015ScalableRW}.
Combining the
ideas of PF and SBM, the infinite Edge Partition Model
(EPM)~\cite{zhou2015infinite} and its extensions~\cite{hu2016topic}
have proven successful for relational networks.


When a network has less data, relational
learning becomes more difficult.
One extreme case is the {\it cold-start}
problem~\cite{Lin:2013,Sedhain:2014,Zhang:2015}, where
a node has no observed links,
making suggestion of links for that node even more challenging.
In such cases, it is natural to appeal to side information
such as node attributes or features.
For instance, papers in citation networks are often associated with
categories and authors, and users in Facebook or Twitter are
often asked to provide information such as age, gender and interests. It
is reasonable to assume that nodes having similar attributes are
more likely to relate to each other (i.e., homophily, \citealp{Nickel:2016jd}). Thus, node attributes serve as important
complementary information to relational data. 

There are few Bayesian probabilistic
relational models that are able to leverage side information. For example, 
NLFM
uses a linear regression model to transform the features of each node into a
single number, which contributes to link probabilities.
However, side information in NLFM
cannot directly influence the latent factors, which gives
little support for community detection. As an extension of MMSB, the
Non-parametric Meta-data Dependent Relational (NMDR)
model~\cite{kim2012nonparametric} incorporates attributes into the
mixed-membership distribution of each node with the logistic-normal
transform, which results in non-conjugacy for
inference. \citet{fan2016learning} further developed this idea in the
Node information Involved Mixture Membership model (niMM), where side
information is integrated in a conjugate way. Although these models
demonstrate improvement using side information, they scale
quadratically in the number of nodes and the incorporation of side
information is often complicated.

Several recent methods~\cite{gopalan2014content,acharya2015gamma,hu2016non}
extend PF with side information
using the additivity of the Poisson and
gamma distributions/processes.
With improved scalability, 
the Structural Side Information Poisson Factorisation
(SSI-PF)~\cite{hu2016non} 
models directed unweighted networks with node labels, such as
citation networks with papers labelled with one of several
categories. However, its performance remains untested 
when a node has multiple attributes.
Moreover, undirected networks are not handled by SSI-PF.

In this paper we present the Node Attribute Relational Model (NARM)\footnote{Code available at \url{https://github.com/ethanhezhao/NARM/}}, a fully
Bayesian approach that models large, sparse, and
unweighted relational networks with arbitrary node attributes encoded in binary
form.  It works with Poisson gamma relational models to
incorporate side information. Specifically, we propose
the Symmetric NARM (Sym-NARM) for undirected networks, an extension of EPM
\cite{zhou2015infinite} and the Asymmetric NARM (Asym-NARM) for
directed networks, an extension of PF \cite{zhou2012beta}. The proposed models have
several key properties: \textbf{(1)~Effectively modelling node attributes:} the
proposed models are able to achieve improved link prediction performance,
especially  where training data are limited. \textbf{(2)~Fully
Bayesian and conjugate:} the inference is done by efficient, closed-form
Gibbs sampling which scales linearly in the number of observed links and takes
advantage of the sparsity of node attributes. It makes our models
scalable for large but sparse relational networks with large sets of node
attributes. \textbf{(3)~Flexibility}: the proposed models work on directed and
undirected relational networks with flat and hierarchical node attributes. 



\begin{figure*}[!t]
     \vspace{-4mm}

\begin{center}
\begin{multicols}{2}
     \begin{IEEEeqnarray}{+rCl+x*}
      \label{eq_sym_bpl1}
      y_{i,j} &=& \textbf{1}_{(x_{i,j} > 0)} \\
      \label{eq-2}
      x_{i,j} &=& \sum_{k_1,k_2 = 1}^{K} x_{i,k_1,k_2,j} \\
      \label{eq_sym_bpl2}
      x_{i,k_1,k_2,j} &\sim& \text{Poi}(\phi_{i,k_1} \lambda_{k_1,k_2} \phi_{j,k_2})\\
      \label{eq_sym_na1}
      \phi_{i,k}  &\sim& \mathrm{Ga}(g_{i,k}, 1/c_{i})\\
      \label{eq-5-exp}
      g_{i,k} &=& b_{k} \prod_{l=1}^{L} h_{l,k}^{f_{i,l}}\\
      \label{eq_sym_na2}
      h_{l,k} &\sim& \mathrm{Ga}\left(\mu_0, 1/ (1/\mu_0)\right)\\
      b_{k} &\sim& \mathrm{Ga}\left(\mu_0, 1/ (1/\mu_0)\right)\\
      \label{eq_sym_hgp1}
      \lambda_{k_1,k_2} &\sim& 
      \begin{dcases}
          \mathrm{Ga}(\epsilon r_{k}, 1/a_0) ,& \text{if } k_1 = k_2 = k\\
          \mathrm{Ga}(r_{k_1} r_{k_2}, 1/a_0),  & \text{otherwise} 
      \end{dcases} \\
      \label{eq_sym_hgp2}
      r_{k} &\sim& \mathrm{Ga}(\gamma_0 / K, 1 / c_0) 
      \end{IEEEeqnarray} 
     \vspace{-10mm}
      \caption{The generative model of Sym-NARM. $\textbf{1}_{(\cdot)}$ is the indicator function. $\text{Poi}(\cdot)$ and
      $\text{Ga}(\cdot,\cdot)$ stand for the Poisson distribution and the gamma
      distribution respectively. Conjugate gamma priors are imposed on the
      hyper-parameters: $\gamma_0$, $\epsilon$, $c_0$, $c_i$, and $a_0$.} 
      \label{fg_sym_narm}\par      
      \begin{IEEEeqnarray}{+rCl+x*}
      y_{i,j} &=& \textbf{1}_{(x_{i,j} > 0)} \\
      x_{i,j} &\sim& \sum_{k}^{K}  x_{i,j,k} \\ 
      \label{eq-11}
      x_{i,j,k} &\sim& \text{Poi}(\phi_{i,k} \theta_{j,k})\\
      \label{eq-12}
      \phi_{i,k}  &\sim& \mathrm{Ga}\left(g_{i,k}, \frac{q_k}{1 - q_k}\right)\\
      \label{eq-13}
      q_k &\sim& \mathrm{Be}\left(c_0 \epsilon, c_0 (1 - \epsilon)\right)\\
      \label{eq_g_d}
      g_{i,k} &=& b_{k}\prod_{l=1}^{L} h_{l,k}^{f_{i,l}}\\
      \label{eq-14}
      h_{l,k} &\sim& \mathrm{Ga}\left(\mu_0, 1/ (1/\mu_0)\right)\\
      b_{k} &\sim& \mathrm{Ga}\left(\mu_0, 1/ (1/\mu_0)\right)\\
      \label{eq-18}
      \theta_{:,k} &\sim& \mathrm{Dir}_N(a_0\vec{1}) 
      \end{IEEEeqnarray} 
     \vspace{-10mm}
      \caption{The generative model of Asym-NARM. $\mathrm{Dir}_N(\cdot)$ and
      $\mathrm{Be}(\cdot,\cdot)$ stand for the $N$ dimensional
      Dirichlet distribution and the beta distribution respectively. $\mu_0, \nu_0,
      a_0, e_0, f_0, c_0, \epsilon$ are the hyper-parameters.}
      \label{fg_asym_narm}\par
\end{multicols}
\end{center}
\label{fg_auc_undirected}
\vspace{-5mm}
\end{figure*}

\section{The Node Attribute Relational Model}
\label{sec-model}
Here we focus on modelling unweighted networks that can be either directed (i.e., the relationship is asymmetric) or undirected.
Assume a relational network with $N$ nodes is stored
in a binary adjacency matrix $\matr{Y} \in \{0,1\}^{N \times N}$ where $y_{i,j} = 1$ indicates the presence of a link between nodes $i$ and $j$.
If the relationship described in the network is symmetric, 
then $y_{i, j} = y_{j,i}$,
and if asymmetric, possibly $y_{i, j} \neq y_{j,i}$.
Node attributes are encoded in a binary matrix $\matr{F} \in
\{0,1\}^{N \times L}$, where $L$ is the total number of attributes. 
Attribute $f_{i,l} = 1$ indicates attribute $l$ is active with node $i$ and vice
versa. Although our models incorporate binary attributes,
categorical attributes and real-valued attributes can be converted into binary values with proper
transformations~\cite{kim2012nonparametric,fan2016learning,hu2016non}. 

\subsection{The Symmetric Node Attribute Relational Model}

Sym-NARM works with undirected networks.
Its generative process is shown in Figure
\ref{fg_sym_narm}.
Instead of modelling the binary matrix $\matr{Y}$ directly, it applies the
Bernoulli-Poisson link (BPL) function~\cite{zhou2015infinite}
using an underlying latent count matrix $\matr{X}$.
One first
draws a latent count $x_{i,j}$ from the Poisson distribution and then
thresholds it at 1 to generate a binary value $y_{i,j}$. This is
shown in Eqs.~(\ref{eq_sym_bpl1})-(\ref{eq_sym_bpl2}). Analysed in
\cite{zhou2015infinite,hu2016topic,hu2016non}, BPL has the appealing property
that if $y_{i,j} = 0$, then $x_{i,j} = 0$ with probability one. Thus, only
non-zeros in $\matr{Y}$ need to be sampled, giving huge
computational savings for large sparse networks, illustrated in Section~\ref{sec-inf} and Section~\ref
{sec-runtime}.

The latent matrix $\matr{X}$ is further factorised into $K$ latent factors
with a non-negative bilinear model: $\matr{X} \sim \mathrm{Poi}(\matr{\Phi}
\matr{\Lambda} \matr{\Phi}^T)$ where $\matr{\Phi} \in \mathbb{R}_{+}^{N
\times K}$ and $\matr{\Lambda} \in \mathbb{R}_{+}^{K \times K}$.
$\matr{\Phi}$ is referred to as the {\em node factor loading matrix} where
$\phi_{i,k}$ models the strength of the connection between node $i$
and latent factor $k$. As in SBM, the
correlations of the latent factors are modelled in a
symmetric matrix $\matr{\Lambda}$,
referred to as the {\em block matrix}. Following
\cite{zhou2015infinite}, we draw $\matr{\Lambda}$ from a
hierarchical relational gamma process (implemented with truncation as a
vector of gamma variables)
, shown in
Eqs.~(\ref{eq_sym_hgp1}) and~(\ref{eq_sym_hgp2}).

One appealing aspect of our model
is the incorporation of node attributes on the prior
of $\phi_{i,k}$ (i.e., $g_{i,k}$). Shown in
Eq.~(\ref{eq-5-exp}), $g_{i,k}$ is constructed with a log linear combination of $f_{i,l}$. $h_{l,k}$ is referred to as the $k^{\text{th}}$ {\em attribute factor loading} of attribute $l$,
which influences $g_{i,k}$ iff attribute $l$ is active with node $i$ (i.e., $f_{i,l} =
1$). $b_{k}$ acts as an attribute-free bias
for each latent factor $k$.
$h_{l,k}$ and $b_{k}$ are
gamma distributed with mean $1$, hence if attribute $l$ does not
contribute to latent factor $k$ or is less useful, $h_{l,k}$ is expected to be
near $1$ and to
have little influence on $g_{i,k}$. 
The hyper-parameter $\mu_0$ controls the variation of $h_{l,k}$.

The intuition of our model is: if two
nodes have more common attributes, their gamma shape parameters
will be more similar, with similar node
factor loadings, resulting in a larger probability that they relate to each other.
Moreover, instead of incorporating the node attributes directly into the node factor loadings,
Sym-NARM uses them as the prior information using
Eq.~(\ref{eq_sym_na1}), which results in a principled way of balancing the side information and the network data.
In addition, different attributes can
contribute differently to the latent factors.
For example, the gender of
an author may be much less important to co-authorship with
others than the research fields. 
This is controlled by the attribute factor loading $h_{l,k}$ in our model.


\subsection{The Asymmetric Node Attribute Relational Model}
Extending the Beta Gamma Gamma Poisson factorisation
(BGGPF)~\cite{zhou2012beta}, Asym-NARM works on directed relational
networks with node attributes incorporated in a similar way to
Sym-NARM. Figure~\ref{fg_asym_narm} shows its generative process. Here
the latent count matrix $\matr{X}$ is factorised as $\matr{X} \sim
\text{Poi}(\matr{\Phi} \matr{\Theta})$, where $\matr{\Phi} \in
\mathbb{R}_{+}^{N \times K}$ and
$\matr{\Theta} \in \mathbb{R}_{+}^{K \times N}$
are referred to as the \emph{factor loading matrix}
and the  \emph{factor score matrix} respectively. Similar to SSI-PF, the node
attributes are incorporated on the prior of $\matr{\Phi}$.


\subsection{Incorporating Hierarchical Node Attributes}
Relational networks can be associated with hierarchical
side information~\cite{hu2016non}. For example, in a patent citation network, patents can be labelled with the International Patent Classification (IPC) code, which is a hierarchy of patent categories and sub-categories.
Suppose the second
level attributes are stored in a binary matrix $\matr{F'} \in \{0,1\}^{L \times
M}$ where $M$ is the number of attributes in the second level. Our models
can be used to incorporate hierarchical node
attributes via a straightforward extension:
replace hyper-parameter  $\mu_0$ in Eq.~(\ref{eq_sym_na2})
with  $\mu_{l,k} = \prod_m^{M} \delta^{f'_{l,m}}_{m,k}$.
This extension mirrors
what is done for first level attributes.

\section{Inference with Gibbs Sampling}
\label{sec-inf}

Both Sym-NARM and Asym-NARM enjoy local conjugacy so the inference of
all latent variables can be done by closed-form Gibbs sampling.
Moreover, the inference only needs to be conducted on the 
non-zero entries in $\matr{Y}$ and $\matr{F}$.
This section focuses on the sampling of $h_{l,k}$ ($b_k$),
the key variable in the proposed incorporation of node attributes. 
The sampling of the other latent variables is similar to
those in EPM and BGGPF, detailed in
\cite{zhou2015infinite,zhou2012beta}. As the sampling for
$h_{l,k}$ is analogous in Sym-NARM and Asym-NARM, our introduction will be based on Asym-NARM alone.



%

\setlength{\belowdisplayskip}{1pt} \setlength{\belowdisplayshortskip}{1pt}
\setlength{\abovedisplayskip}{1pt} \setlength{\abovedisplayshortskip}{1pt}
With the Poisson gamma conjugacy, the likelihood for $g_{i,k}$ with $\phi_{i,k}$ marginalised out is:
\begin{IEEEeqnarray}{+rCl+x*}
  \prob{g_{i,k} | x_{i,\cdot,k}} &\propto& (1-q_k)^{g_{i,k}}
  \frac{\Gamma(g_{i,k} + x_{i,\cdot,k})}{\Gamma(g_{i,k})}
\label{eq_p_g}
\end{IEEEeqnarray}
where $x_{i, \cdot, k} = \sum_{j} x_{i,j,k}$ and $x_{i,j,k}$ is the latent count.
The gamma ratio in Eq.~(\ref{eq_p_g}), i.e., the Pochhammer symbol for a rising factorial,
can be augmented 
with an auxiliary variable $t_{i,k}$:
$\frac{\Gamma(g_{i,k} + x_{i,\cdot,k})}{\Gamma(g_{i,k})} = \sum_{t_{i,k}=0}^{x_{i,\cdot,k}} S^{x_{i,\cdot,k}}_{t_{i,k}} g_{i,k}^{t_{i,k}}$
where $S^{x}_{t}$ indicates an unsigned Stirling number of the first kind~\cite{chen2011sampling,teh2012hierarchical,zhou2015negative}.

Taking $\mathcal{O}(x_{i,\cdot,k})$, $t_{i,k}$ can be directly sampled by a Chinese Restaurant Process with $g_{i,k}$ as the concentration and $x_{i,\cdot,k}$ as the number of customers:
\begin{IEEEeqnarray}{+rCl+x*}
t_{i,k} &\leftarrow& t_{i,k} + \text{Bern}\left(\frac{g_{i,k}}{g_{i,k}+i'}\right)
~\hfill \text{for~} i' = 1:x_{i,\cdot,k}
\end{IEEEeqnarray} 
where $\text{Bern}(\cdot)$ is the Bernoulli
distribution.
Alternatively, for large $x_{i,\cdot,k}$, because the standard deviation of
$t_{i,k}$ is $\mathcal{O}(\sqrt{\log x_{i,\cdot,k}})$ \cite{BunHut:12},
one can sample $t_{i,k}$ in a small window around the current value
\cite{Du:2010:STM:1842816.1842821}.



With the above augmentation and Eq.~(\ref{eq_g_d}), we get:
\begin{IEEEeqnarray}{+rCl+x*}
\label{eq-20}
\lefteqn{\prob*{\matr{G}, \matr{H} |  x_{:,\cdot,:} , \matr{T}, \matr{F}} ~\propto}~&  \\
&&\prod_{i=1}^N \prod_{k=1}^{K} S^{x_{i,\cdot,k}}_{t_{i,k}} e^{-\log\left(\frac{1}{1 - q_k}\right) g_{i,k}} \cdot \prod_{l=1}^{L} \prod_{k=1}^{K} h_{l,k} ^{\sum_{i=1}^{N} f_{i,l} t_{i,k}}\nonumber
\end{IEEEeqnarray}

Recall that all the attributes are binary and $h_{l,k}$ influences $g_{i,k}$
only when $f_{i,l}=1$. Extracting all the terms related to $h_{l,k}$ in Eq.~(\ref{eq-20}), we get the likelihood of $h_{l,k}$:
\begin{IEEEeqnarray}{+rCl+x*}
\lefteqn{\prob*{ h_{l,k} | \frac{g_{i,k}}{h_{l,k}}, t_{:,k}, f_{:,l}} ~\propto}&& \\
&&e^{- h_{l,k} \log\left(\frac{1}{1 - q_k}\right) \sum_{i=1:f_{i,l}=1}^{N} \frac{g_{i,k}}{h_{l,k}}} h_{l,k} ^{\sum_{i=1}^{N} f_{i,l} t_{i,k}}  \nonumber
\end{IEEEeqnarray}
where $\frac{g_{i,k}}{h_{l,k}}$ is the value of $g_{i,k}$ with $h_{l,k}$ removed when $f_{i,l}=1$. 
The likelihood function above is in a form that is conjugate to the gamma prior.
Therefore, it is straightforward to yield the following sampling strategy for $h_{l,k}$:
\setlength{\belowdisplayskip}{9pt} \setlength{\belowdisplayshortskip}{9pt}
\begin{IEEEeqnarray}{+rCl+x*}
\label{eq_delta_1}
h_{l,k} &\sim& \mathrm{Ga}( \mu', 1/\nu')\\
\label{eq_delta_2}
\mu' &=& \mu_0 +  \sum_{i=1: f_{i,l} = 1}^{N}  t_{i,k} \\
\label{eq_delta_3}
\nu' &=& 1/\mu_0 - \log\left(1 - q_k\right) \sum_{i=1:f_{i,l}=1}^{N}  \frac{g_{i,k}}{h_{l,k}}
\vspace{-5mm}
\end{IEEEeqnarray}

Precomputed with Eq.~(\ref{eq_g_d}), $g_{i,k}$ can be updated with Eq.~(\ref{eq_g}), after $h_{l,k}$ is sampled.
\setlength{\belowdisplayskip}{1pt} \setlength{\belowdisplayshortskip}{1pt}
\begin{IEEEeqnarray}{+rCl+x*}
\label{eq_g}
g_{i,k} \leftarrow \frac{g_{i,k} h'_{l,k}}{h_{l,k}} 
~\hfill \text{for~} i = 1:N \text{~and~} f_{i,l} = 1
\end{IEEEeqnarray}
where $h'_{i,k}$ is the newly sampled value of $h_{i,k}$.
 
To compute Eqs.~(\ref{eq_delta_2})-(\ref{eq_g}), we only need to
iterate over the nodes that attribute $l$ is active with (i.e., $f_{i,l}=1$).
Thus, the sampling for
$\matr{H}$ takes $\mathcal{O}(D'KL)$ where $D'$ is the average number
of nodes that an attribute is active with. This demonstrates how the sparsity of node
attributes is leveraged. As the mean of $x_{i,\cdot,k}$ is $D/K$, sampling the tables $\matr{T} \in \mathbb{N}^{N \times K}$ takes $\mathcal{O}(ND)$ which can be accelerated with the window sampling technique explained above.

We show the computational complexity of our and related models in Table~\ref{tb_cc}. The empirical comparison of running speed is in Section~\ref{sec-runtime}. By
taking advantage of both network sparsity and node attribute
sparsity, our models are more efficient than the competitors, especially on large sparse networks with
large sets of attributes.

\section{Related work}
\label{rw}
Compared with the node-attribute models such as 
NMDR and niMM whose methods
result in complicated inference, our
Sym-NARM is much more efficient on large sparse networks, illustrated in Table~\ref{tb_cc}.

\begin{table}[!th]
\centering
\caption{The computational complexity for the compared models. $N$: number of nodes. $K$: number of latent factors. $L$:
      number of node attributes. $D$: the average degree (number of
      edges) per node ($D \ll N$ in sparse networks). $D'$: the
      average number of nodes that an attribute is active with (usually, $D' < N$).
      For the models that incorporate node attributes (marked with a *), the complexity with one level attributes is shown.}
            \vspace{2mm}

\label{tb_cc}
{
\begin{tabular}{|c|c|}
\hline
Model               & \specialcell{Complexity} \\ \hline
\multicolumn{2}{|c|}{Models with the block matrix} \\ \hline
*NMDR~\cite{kim2012nonparametric}              &  $\mathcal{O}(N^2K + NKL)$ \\ \hline
*niMM~\cite{fan2016learning}              &  $\mathcal{O}(N^2K^2 + NKL)$ \\ \hline
EPM~\cite{zhou2015infinite}                &  $\mathcal{O}(NK^2D)$ \\ \hline
*\textbf{Sym-NARM}            &  $\mathcal{O}(NK^2D + D'KL)$ \\ \hline
\multicolumn{2}{|c|}{Models without the block matrix} \\ \hline
BGGPF~\cite{zhou2012beta} &  $\mathcal{O}(NKD)$ \\ \hline
*SSI-PF~\cite{hu2016non}   &  $\mathcal{O}(NKDL)$ \\ \hline
*\textbf{Asym-NARM}           &  $\mathcal{O}(NKD + D'KL)$ \\ \hline
\end{tabular}
  \vspace*{-8mm}
}
\vspace*{-10mm}
\end{table}

The most closely related model to our Asym-NARM,
also extending the BGGPF algorithm,
is SSI-PF.  But it uses the gamma additivity to construct the prior of node factor loadings with the sum of attribute factor loadings.
Our model has several advantages over SSI-PF: (1) The
derivation of Gibbs sampling of SSI-PF requires that each
column of $\matr{\Theta}$ is normalised (Eq.~(\ref{eq-18})). This limits the application of SSI-PF
to other models such as EPM which is an unnormalised model. (2) Shown in
Table~\ref{tb_cc}, Asym-NARM enjoys more efficient computational complexity. 
(3) Shown in Section~\ref{exp}, our model
is more effective especially when a node has
multiple attributes.

There are also models that extend PF and collective matrix factorisation~\cite{singh2008relational} to jointly factorise relational networks and document-word matrices such as \cite{gopalan2014content,Zhang:2015,acharya2015gamma}. 
Our NARM models incorporate general node attributes (not only texts) as the priors of the factor loading matrix in a supervised manner, 
rather than jointly modelling the side information in an
unsupervised manner.

Another related area is supervised topic models such as
\cite{mcauliffe2008supervised,ramage2009labeled,LimBuntine2016}. 
The Dirichlet Multinomial Regression (DMR) model~\cite{mimno2012topic} is the most related one to ours.  It models document attributes on the 
priors of the topic proportions with the logistic-normal
transform. For comparison, we propose DMR-MMSB, extending MMSB with the DMR technique to incorporate side information on the mixed-membership distribution of each node.




\begin{figure*}[!t]
        \centering
         \begin{subfigure}[b]{0.23\linewidth}
                 \centering
                 \caption{Lazega-cowork}
                 \includegraphics[width=1.0\textwidth]{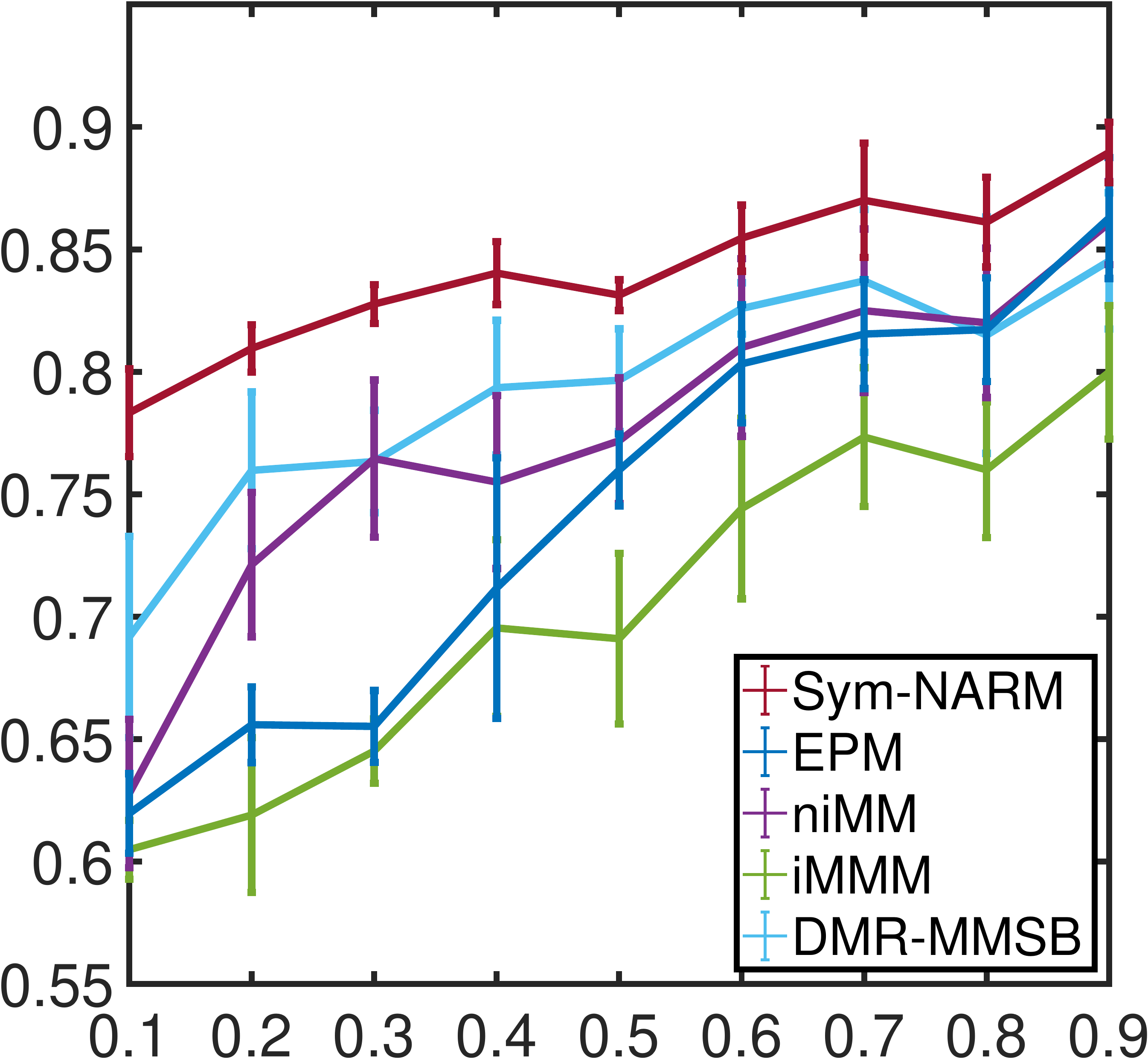}
         \end{subfigure}
         \begin{subfigure}[b]{0.23\linewidth}
                 \centering
                 \caption{NIPS234}

                 \includegraphics[width=0.98\textwidth]{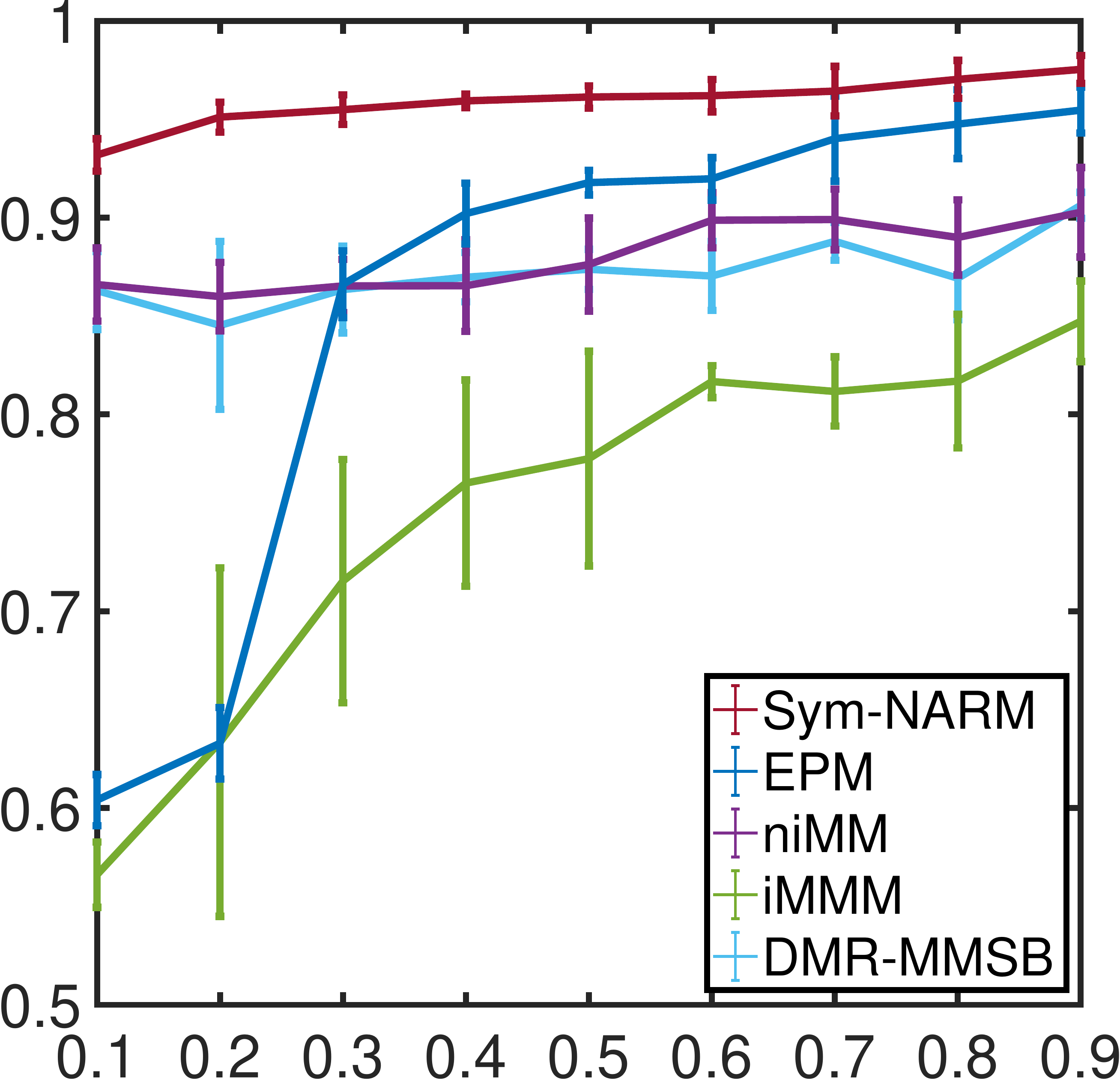}
         \end{subfigure}%
         \begin{subfigure}[b]{0.23\linewidth}
                 \centering
                 \caption{Facebook-ego}
                 \includegraphics[width=1.0\textwidth]{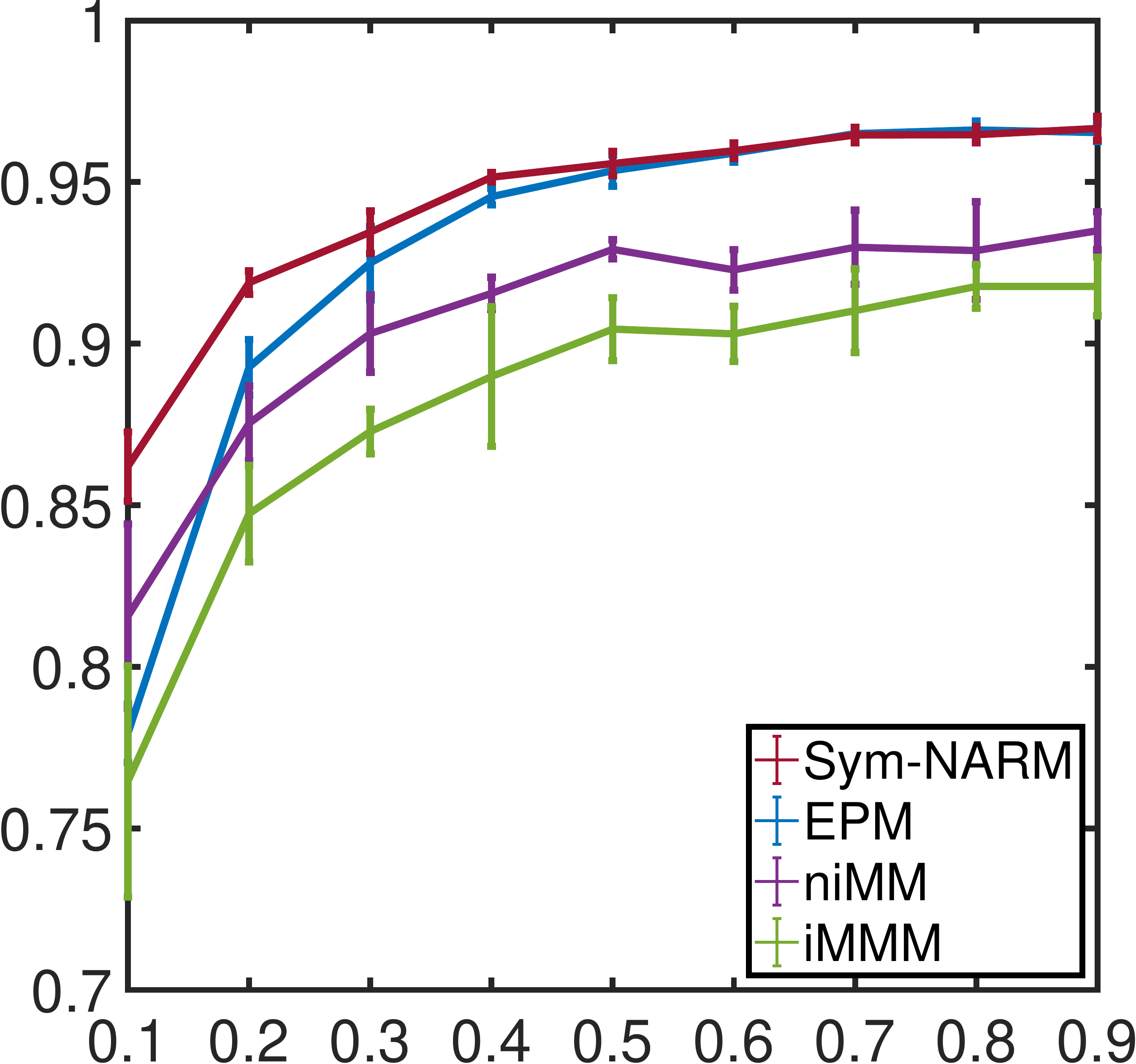}
                 
         \end{subfigure}
         \begin{subfigure}[b]{0.23\linewidth}
                 \centering
                 \caption{NIPS12}
                 \includegraphics[width=1.0\textwidth]{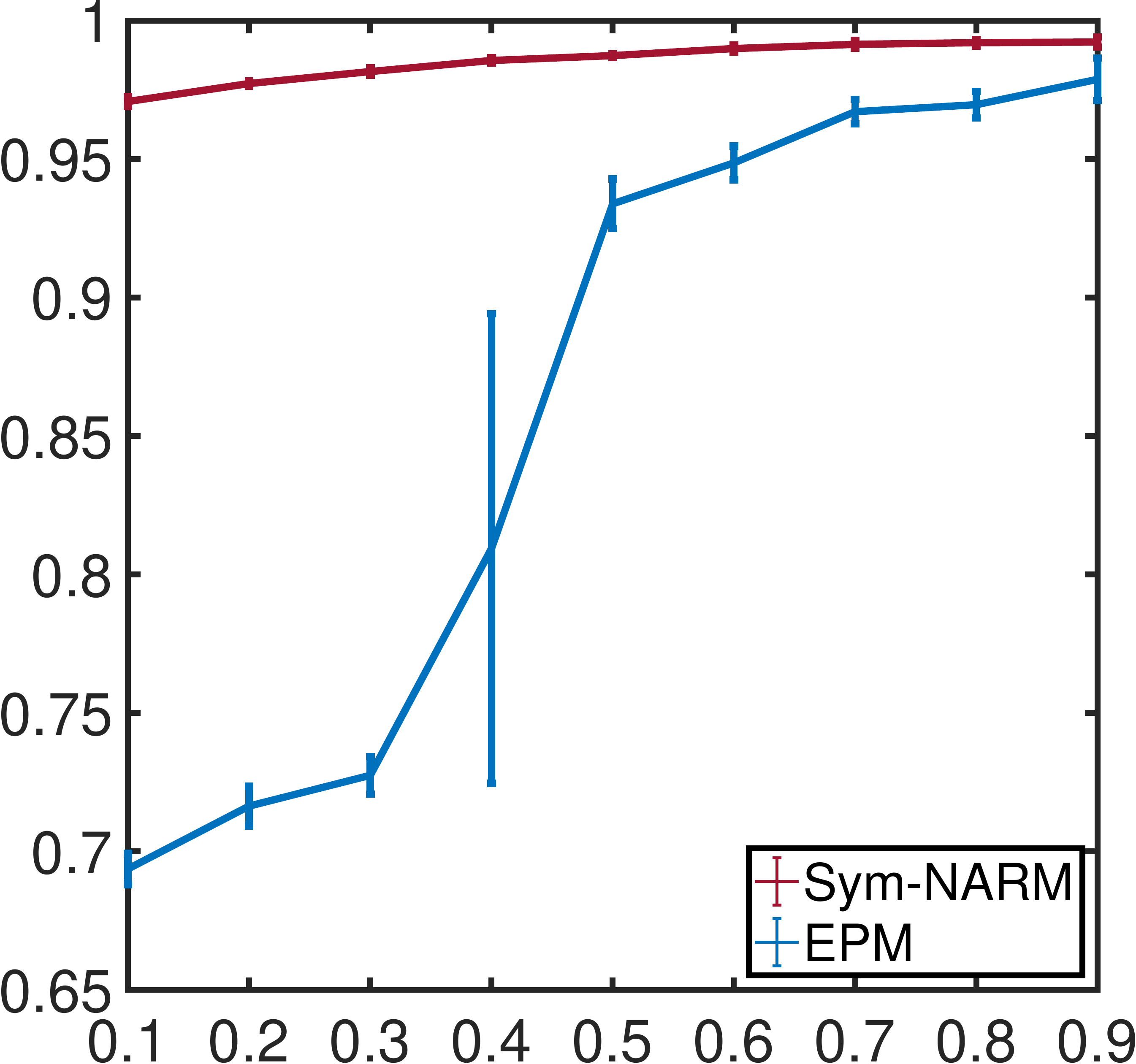}
                 
         \end{subfigure}
        \begin{subfigure}[b]{0.23\linewidth}
                 \centering
                 \includegraphics[width=1.0\textwidth]{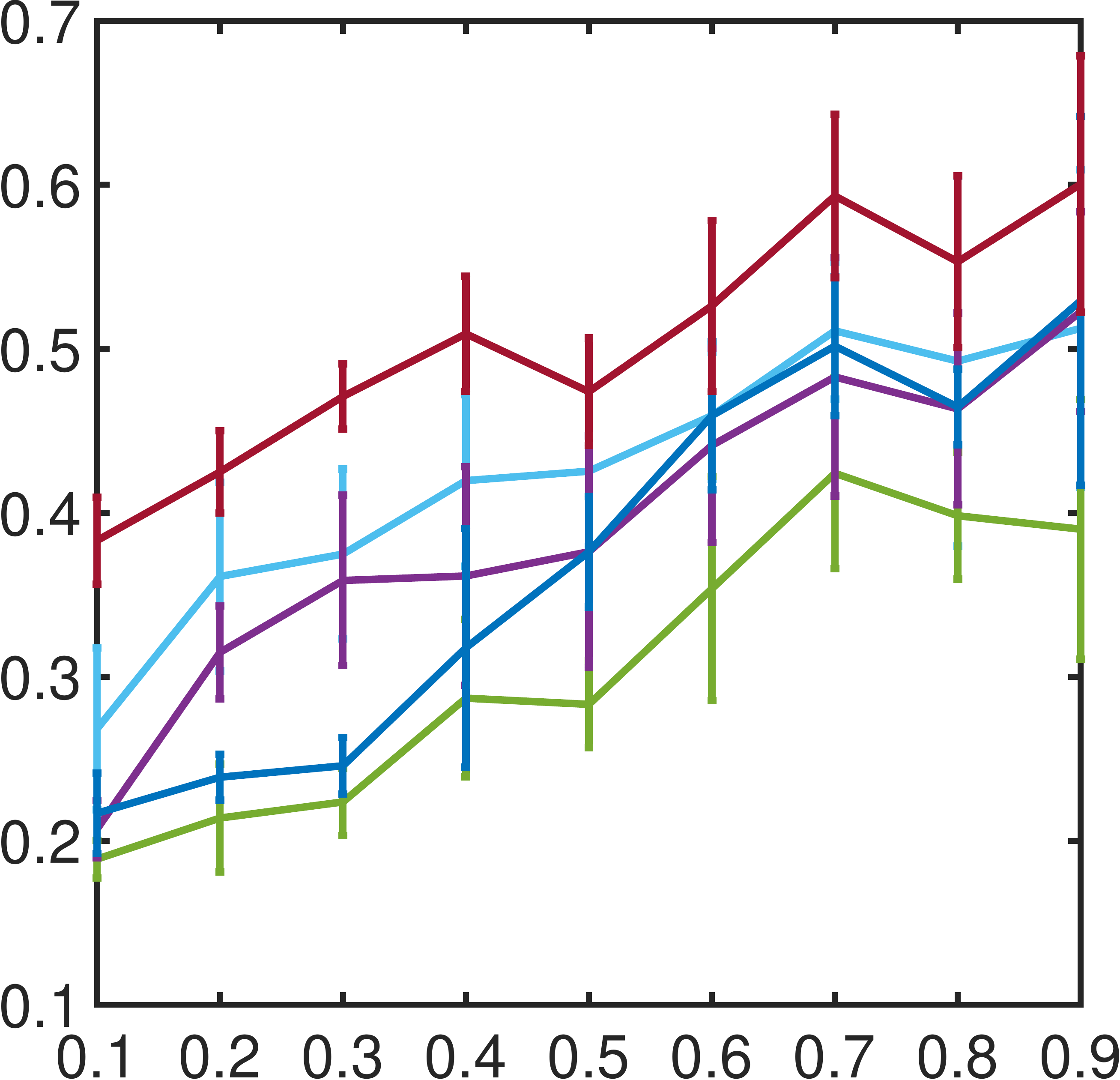}
         \end{subfigure}
         \begin{subfigure}[b]{0.23\linewidth}
                 \centering
                 \includegraphics[width=1.0\textwidth]{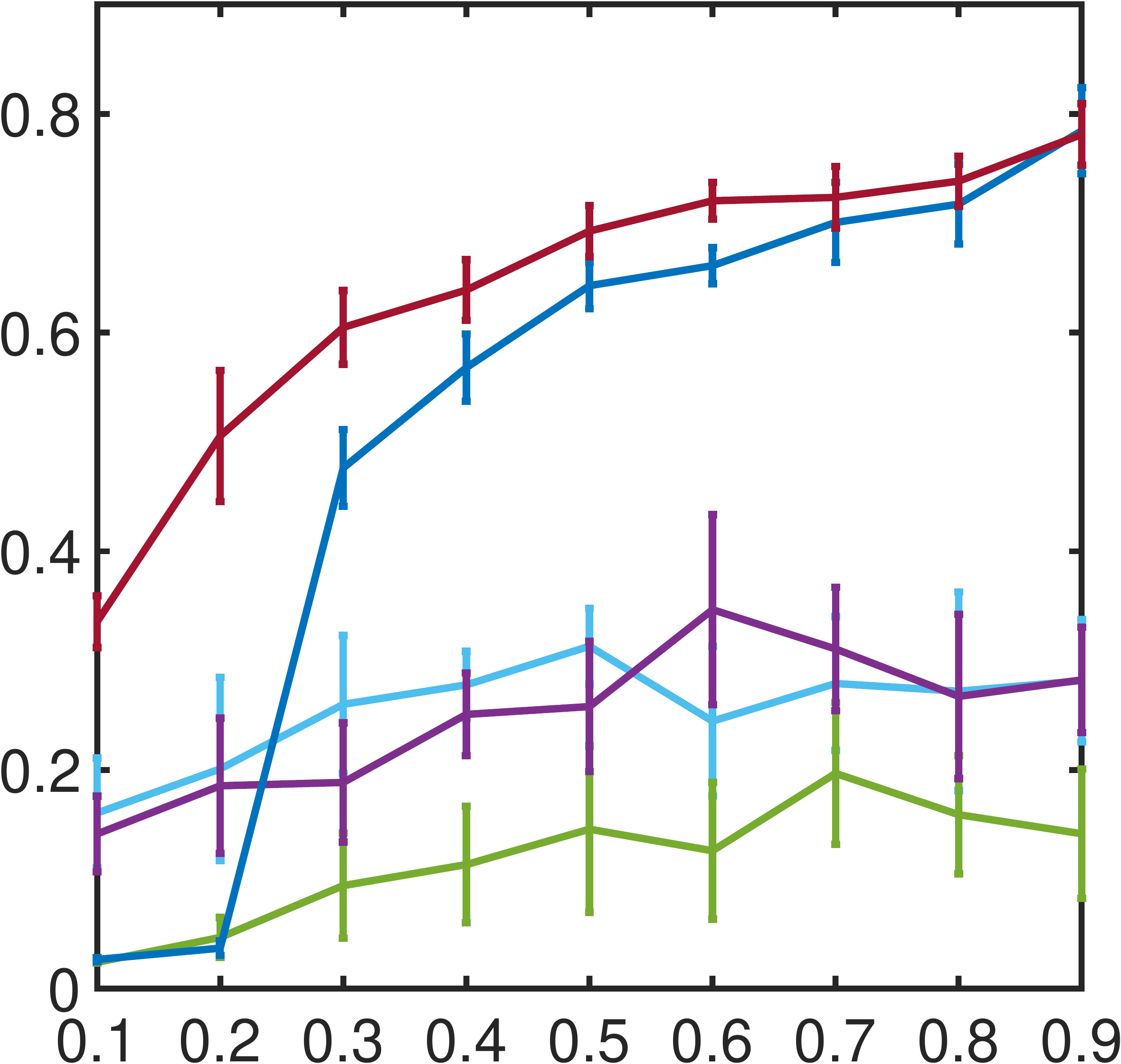}
         \end{subfigure}%
         \begin{subfigure}[b]{0.23\linewidth}
                 \centering
                 \includegraphics[width=1.0\textwidth]{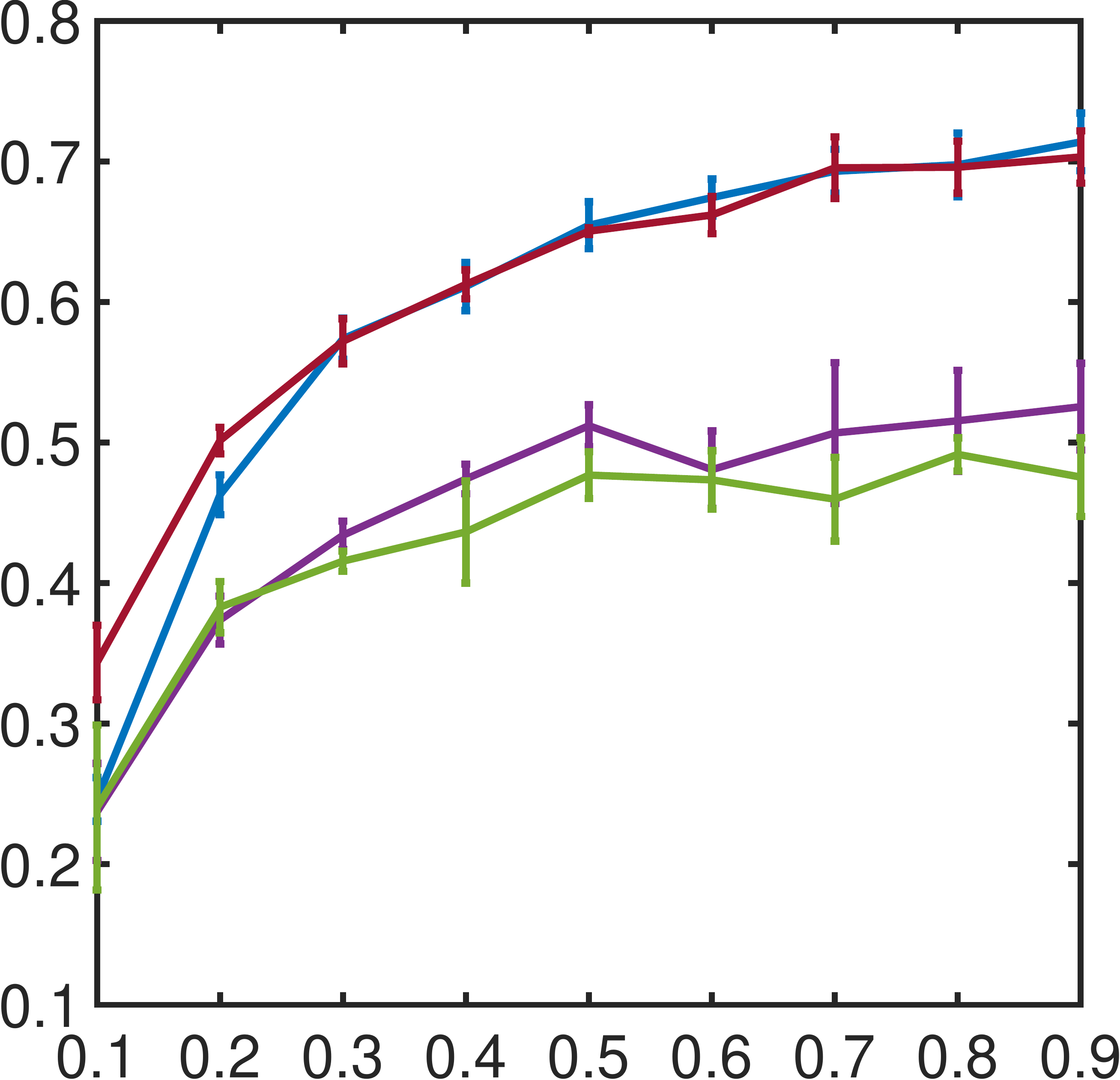}
         \end{subfigure}
         \begin{subfigure}[b]{0.23\linewidth}
                 \centering
                 \includegraphics[width=1.0\textwidth]{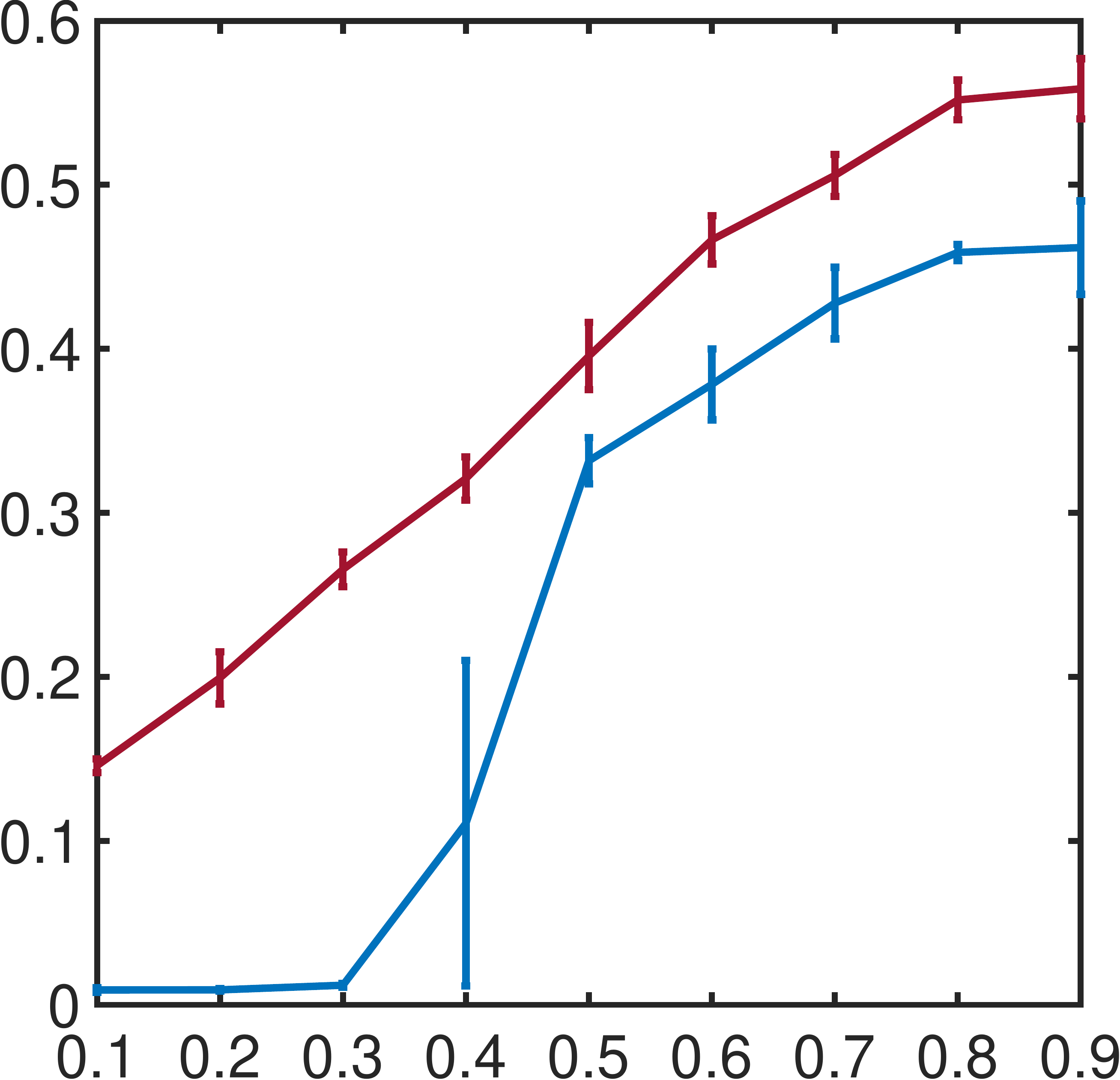}
         \end{subfigure}   
\caption{The AUC-ROC (the first row) and AUC-PR (the second row) scores on the undirected networks. The values on the horizontal axis are the proportions of the training data and each of the error bars is the standard deviation over the five random splits for one proportion. DMR-MMSB achieves its best performance at $K = 5$ and $10$ on Lazega-cowork and NIPS234 respectively.}
\label{fg_auc_pr_undirected}
\vspace{1mm}

        \centering
         \begin{subfigure}[b]{0.23\linewidth}
                 \centering
                  \caption{NIPS234 network}
                 \includegraphics[width=0.98\textwidth]{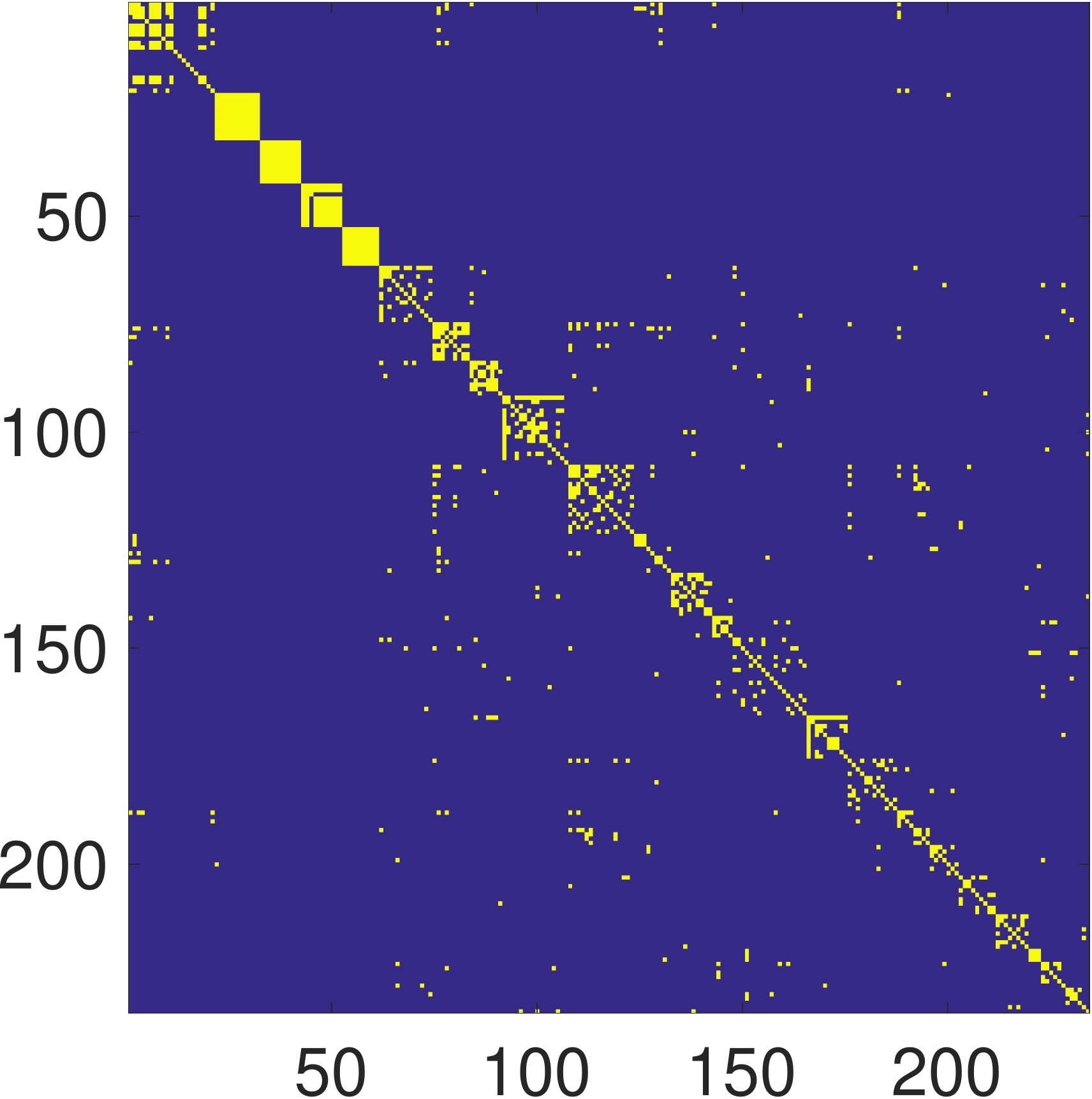}
                 
         \end{subfigure}
         \begin{subfigure}[b]{0.23\linewidth}
                 \centering
                 \caption{Sym-NARM (20\%)}
                 \includegraphics[width=0.98\textwidth]{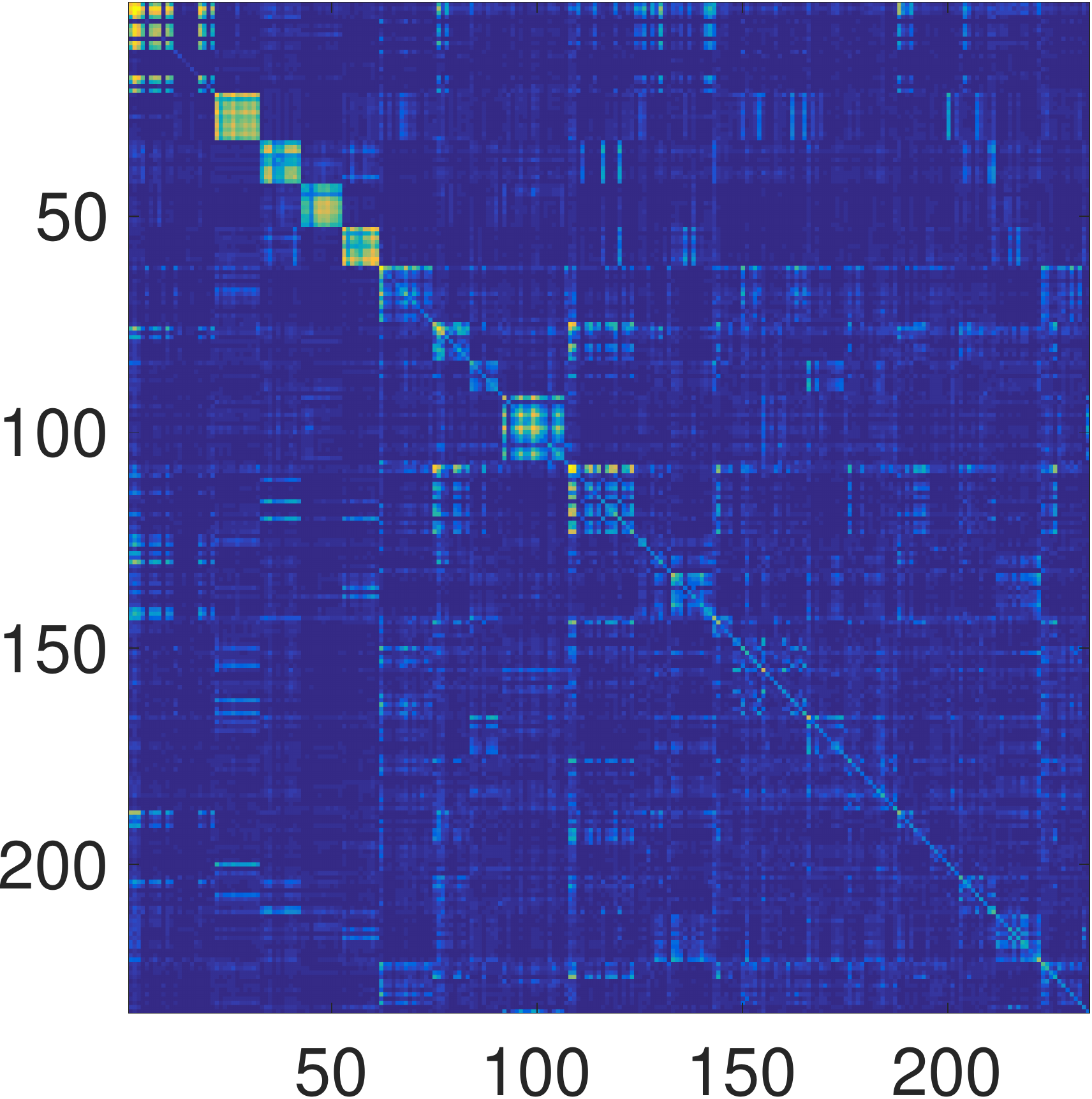}
                
         \end{subfigure}%
         \begin{subfigure}[b]{0.23\linewidth}
                 \centering
                 \caption{EPM (20\%)}
                 \includegraphics[width=0.98\textwidth]{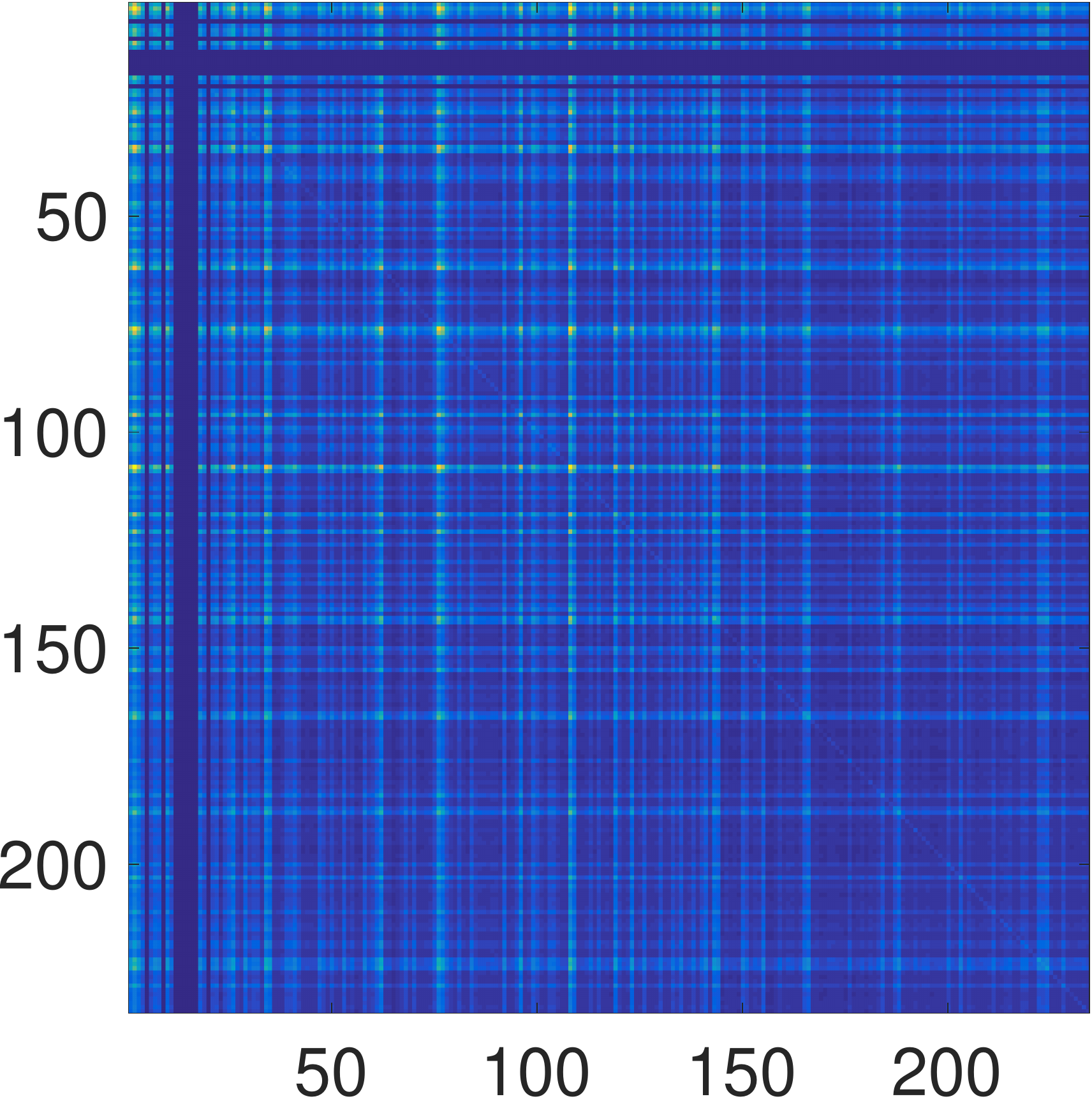}
                  
         \end{subfigure}
         \begin{subfigure}[b]{0.23\linewidth}
                 \centering
                 \caption{niMM (20\%)}
                 \includegraphics[width=0.98\textwidth]{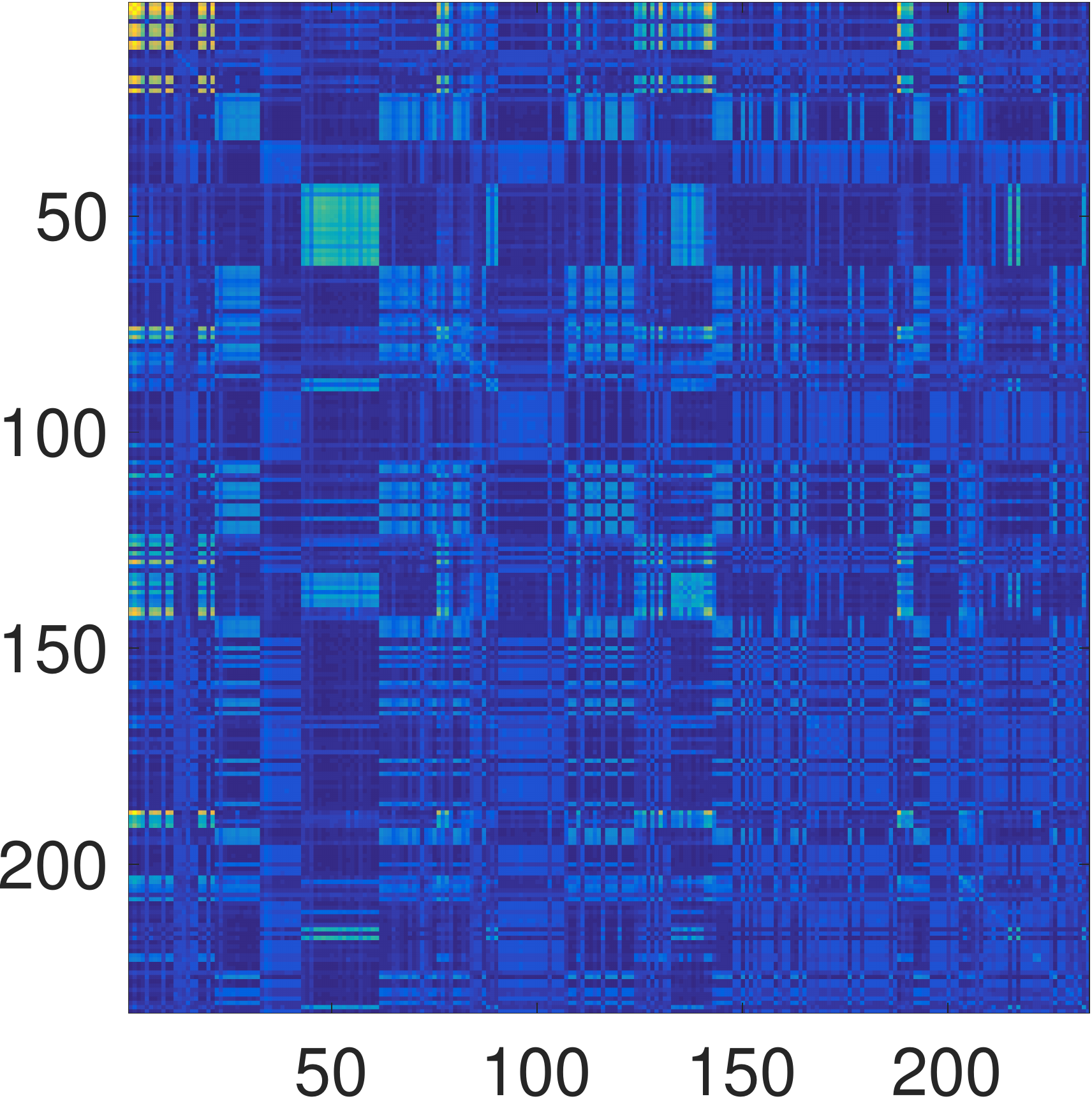}
                 
         \end{subfigure}

         \begin{subfigure}[b]{0.23\linewidth}
                 \centering
                 \caption{Author topic similarity}
                                  \label{atc}
                 \includegraphics[width=0.98\textwidth]{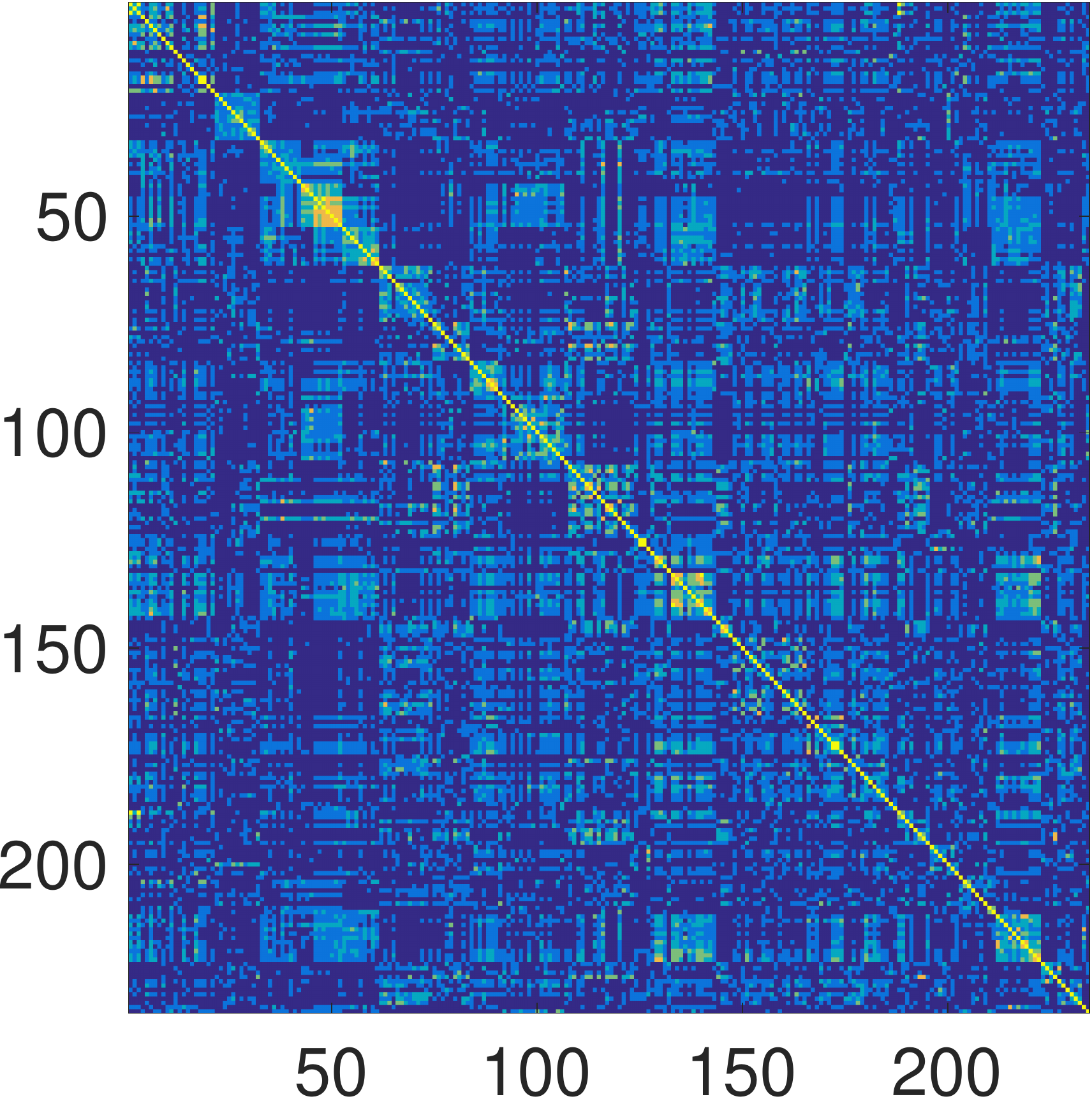}
                 
         \end{subfigure}
         \begin{subfigure}[b]{0.23\linewidth}
                 \centering
                 \caption{Sym-NARM (80\%)}
                 \includegraphics[width=0.98\textwidth]{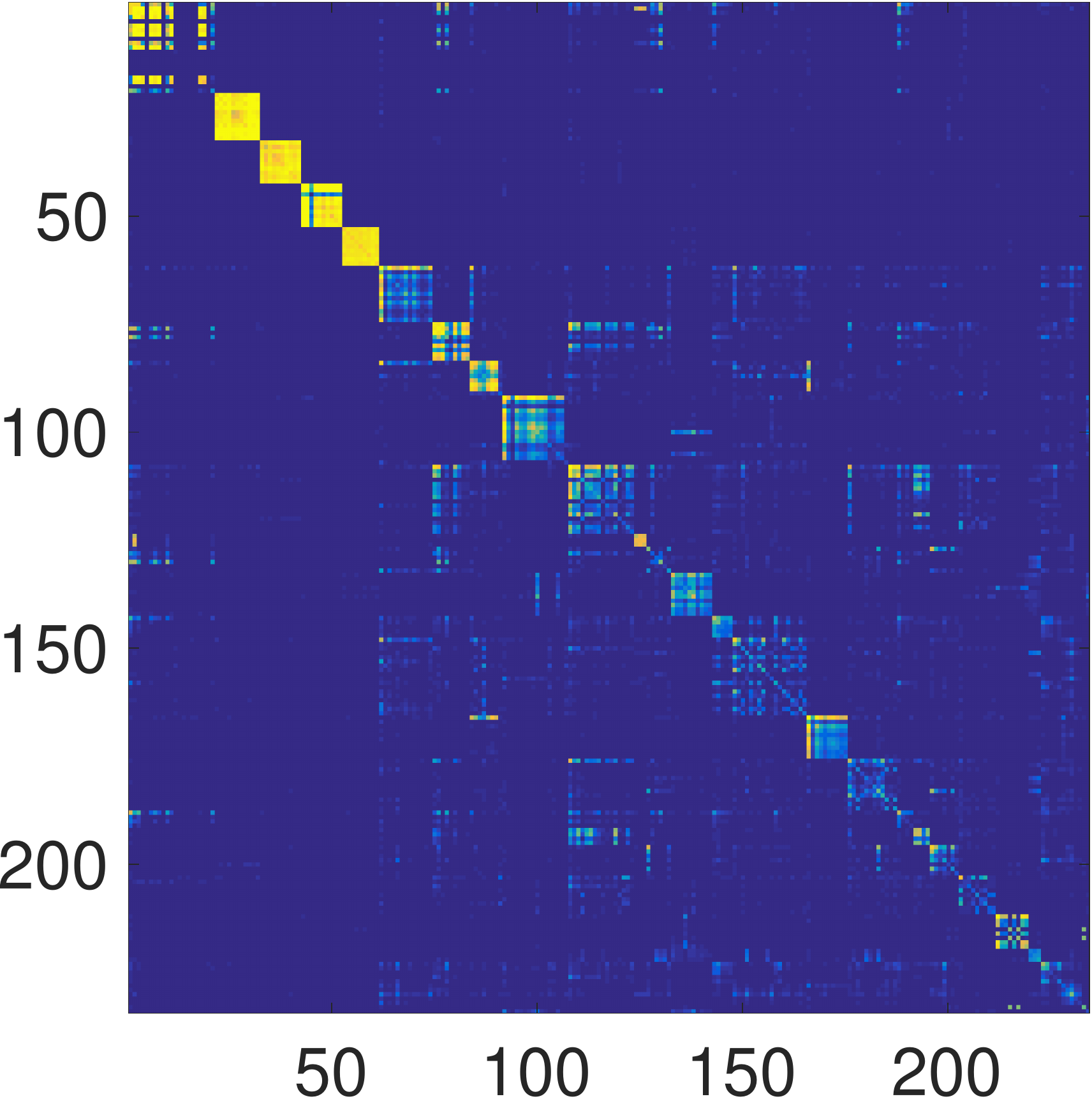}
                 
         \end{subfigure}%
         \begin{subfigure}[b]{0.23\linewidth}
                 \centering
                 \caption{EPM (80\%)}
                 \includegraphics[width=0.98\textwidth]{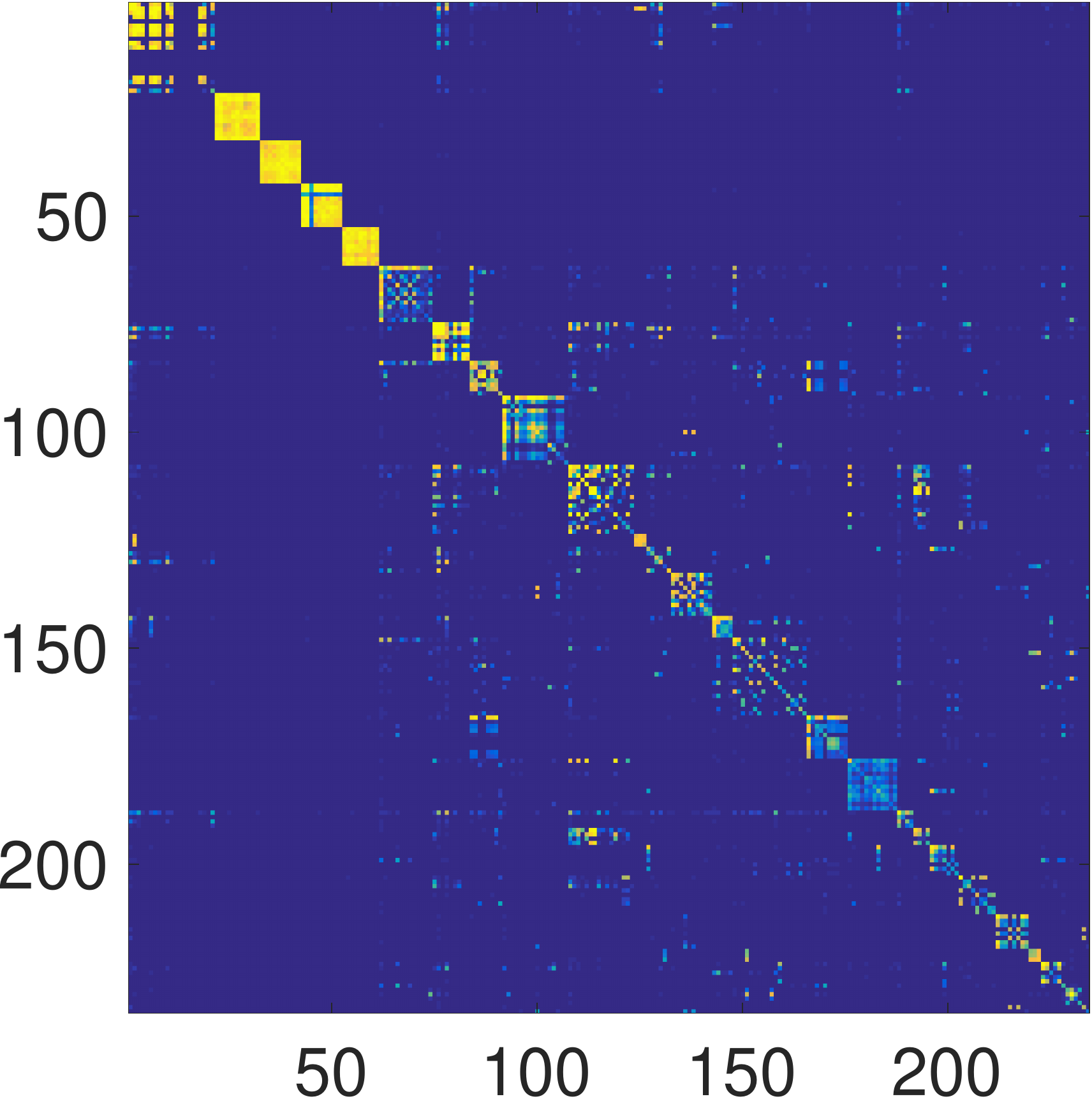}
                 
         \end{subfigure}
         \begin{subfigure}[b]{0.23\linewidth}
                 \centering
                  \caption{niMM (80\%)}
                 \includegraphics[width=0.98\textwidth]{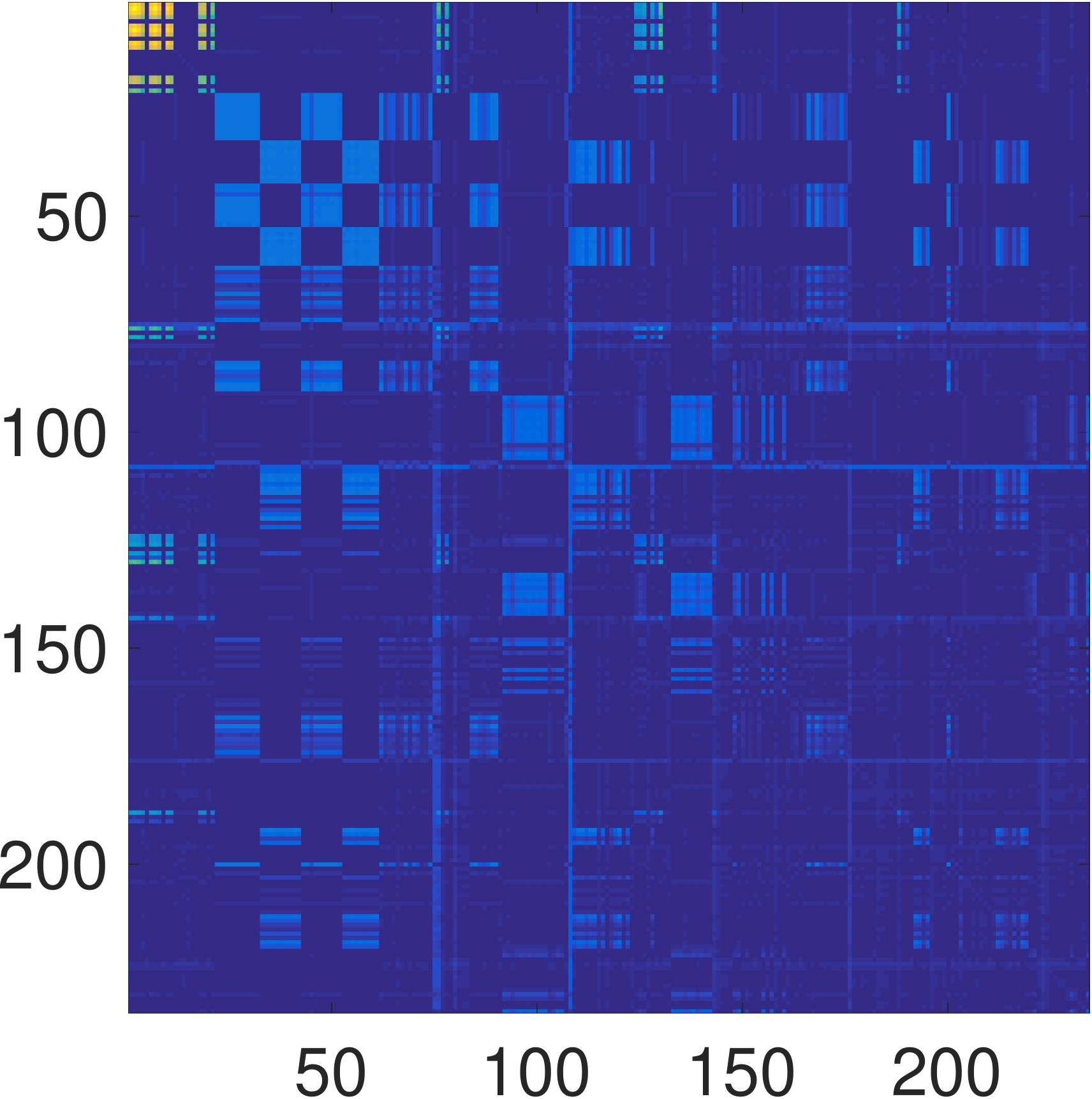}
                
         \end{subfigure}
\caption{The link probability estimations in NIPS234.
Similar to \cite{zhou2015infinite}, the nodes are reordered to make a node with a larger index belong to the same or a smaller-size community, where the disjoint community assignments are obtained by analysing the results of Sym-NARM. (a) The original NIPS234 network. (e) The topic similarity of the authors, obtained by the pairwise cosine distances of the topic proportions, with a brighter colour representing a closer distance. (b)-(d) and (f)-(h) Estimated link probabilities with 20\% and 80\% training data respectively for each compared model.}
\label{fg_reconstruct}
\vspace{-5mm}
\end{figure*}

\section{Experiments}
\label{exp}
In this section we evaluate Sym-NARM and Asym-NARM with a
set of the link prediction tasks on 10 real-world relational datasets
with different sizes and various kinds of node
attributes. We compare our models with the state-of-the-art relational models,
demonstrating that our models outperform the competitors on those
datasets in terms of link prediction performance and per-iteration
running time. We report the average area under the curve of both the
receiver operating characteristic (AUC-ROC) and precision recall 
(AUC-PR) for quantitatively analysing the models. Moreover, we perform qualitative analysis by comparing the link probabilities estimated by the compared models.  

\subsection{Link Prediction on Undirected Networks}
\label{undirected}
For the link prediction 
task on undirected network data,
we compared our \textbf{Sym-NARM} with two models that do not consider node attributes,
\textbf{EPM}~\cite{zhou2015infinite}, a state-of-the-art relational model, and
\textbf{iMMM}~\cite{koutsourelakis2008finding}, a non-parametric version of MMSB,
and two node attribute models,
\textbf{niMM}~\cite{fan2016learning}, a non-parametric relational model which has been
demonstrated to outperform NMDR~\cite{kim2012nonparametric}, and
\textbf{DMR-MMSB}, our extension to MMSB using the Dirichlet Multinomial Regression~\cite{mimno2012topic}.
Sym-NAMR was implemented in MATLAB on top of the EPM code and
we used the code released by the original authors for EPM and niMM. iMMM was implemented by \citet{fan2016learning}
as a variant of niMM.

%

The description of the four datasets used is given below:
\begin{itemize}[noitemsep,leftmargin=4mm,topsep=0pt]
\itemsep0em
\item 
\textbf{Lazega-cowork:} This dataset~\cite{lazega2001collegial} 
contains 378 links of the co-work relationship among 71 attorneys. Each
attorney is associated with attributes such as gender, office
location, and age. After discretisation
and binarisation, we derived a $71 \times 18$ binary node attribute matrix
with 497 non-zero entries.

\item 
\textbf{NIPS234:} 
This is a co-author network of the 234 authors with 598 links
extracted from NIPS 1-17 conferences
\cite{zhou2015infinite}.
We merged all the papers written by the same author as a document, and then trained
a LDA model with 100 topics. The 5 most frequent topics were used as the attributes, which gives us a $234 \times 100$ attribute
matrix with 1170 non-zero entries.
\item 
\textbf{Facebook-ego:} The original dataset \cite{mcauley2012learning}
was collected from survey participants of Facebook users. Out of the
10 circles (i.e., friend lists), we used the first circle that
contains 347 users with 2519 links. Each user is associated with 227
binary attributes, encoding side information such as age, gender, and
education. 
We got a $347 \times 227$ binary node attribute matrix
with 3318 non-zero entries.
\item 
\textbf{NIPS12:} NIPS12 was collected from NIPS papers in vols
0-12. It is a median-size co-author network with 2037 authors
and 3134 links. Similar to NIPS234, we used the 5 most frequent topics as the
attributes for each author. We got a $2037 \times 100$ binary node attribute matrix
with 10185 non-zero entries.
\end{itemize}




\subsubsection{Experimental Settings} 
For each dataset, we
varied the training data from 10\% to 90\% and used the remaining in testing. 
For each proportion, to generate five random splits, we used the code in
the EPM package \cite{zhou2015infinite} which splits a network in terms of its nodes. 
The reported AUC-ROC/PR scores were averaged over the five splits.
We used the default hyper-parameter settings enclosed in the released code
for EPM, niMM and iMMM. 
For our Sym-NARM, we set $\mu_0 = 1$ and all the other
hyper-parameters the same as those in EPM. Note that the models in comparison
except DMR-MMSB are non-parametric models. For Sym-NARM and EPM, we set the truncation level large enough for each dataset: $K_{max} = 50, 100, 256$ for 
Lazega-cowork, Facebook-ego and NIPS234, NIPS12 respectively.  For DMR-MMSB, we varied $K$ in $\{5, 10, 25, 50\}$ and reported the best one.
Following \cite{zhou2015infinite}, we used 3000 MCMC iterations and computed
AUC-ROC/PR with the average probability over the last 1500. The
performance of iMMM and niMM on NIPS12 and  
DMR-MMSB on Facebook-ego and NIPS12 are not reported as the datasets are too large
for them given our computational resources. 

\begin{figure*}[t!p]
        \centering
         \begin{subfigure}[b]{0.23\linewidth}
                 \centering
                 \caption{Lazega-advice}
                                  \label{lazega_advice}

                 \includegraphics[width=1\textwidth]{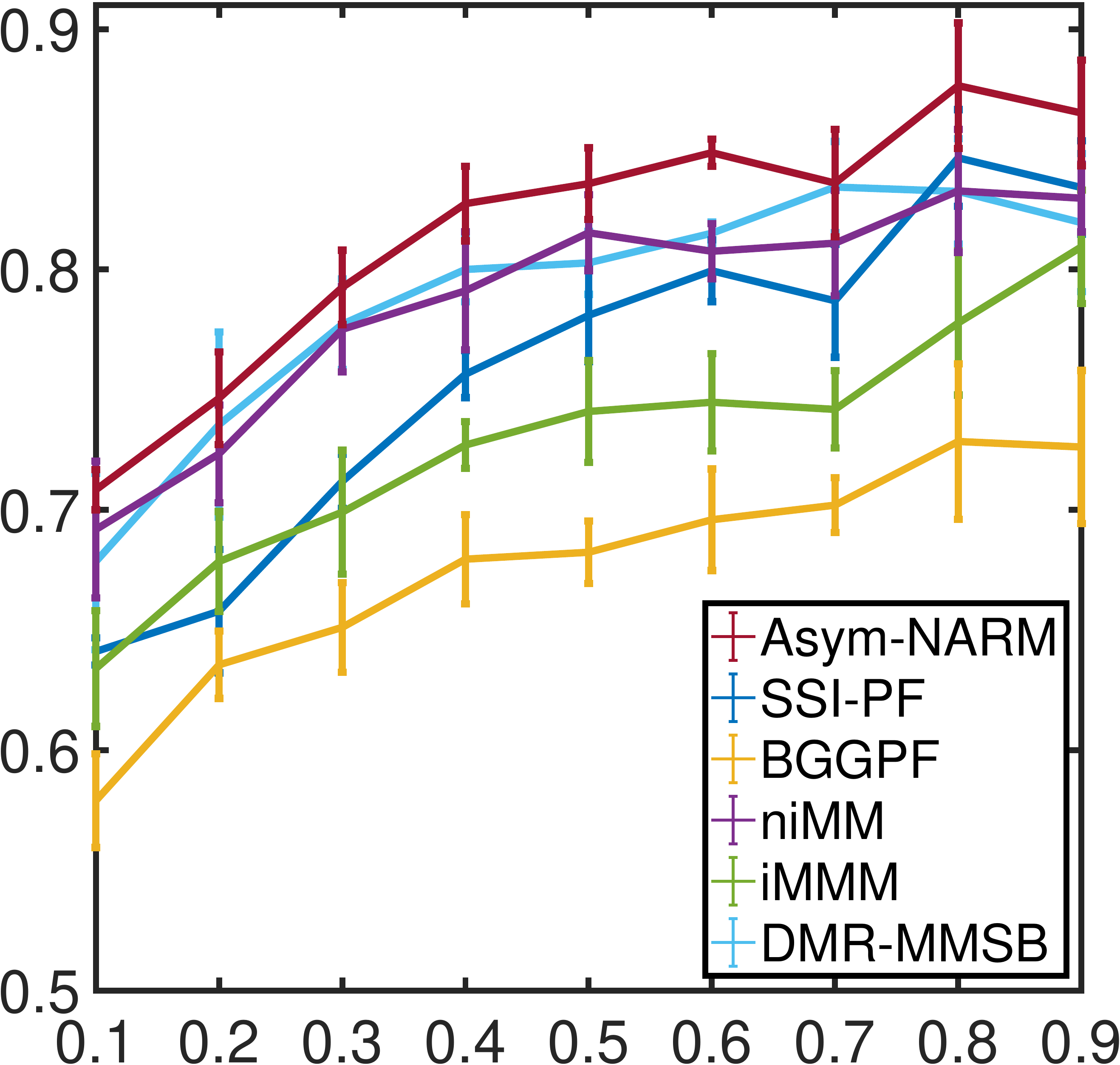}
         \end{subfigure}
         \begin{subfigure}[b]{0.23\linewidth}
                 \centering
                 \caption{Citeseer}
                  \label{citeseer}
                 \includegraphics[width=0.98\textwidth]{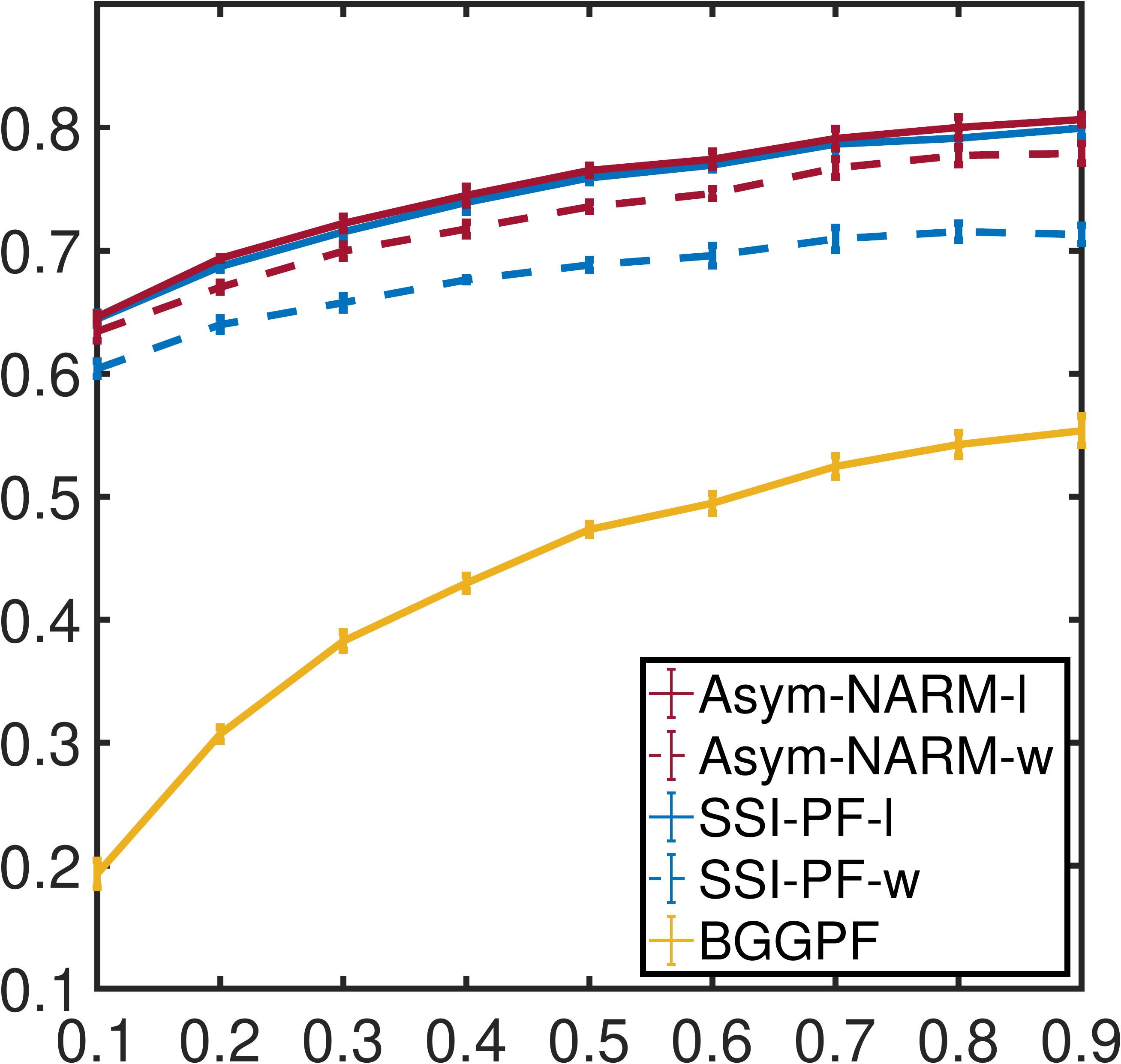}
         \end{subfigure}%
         \begin{subfigure}[b]{0.23\linewidth}
                 \centering
                \caption{Cora}
                \label{cora}
                 \includegraphics[width=0.98\textwidth]{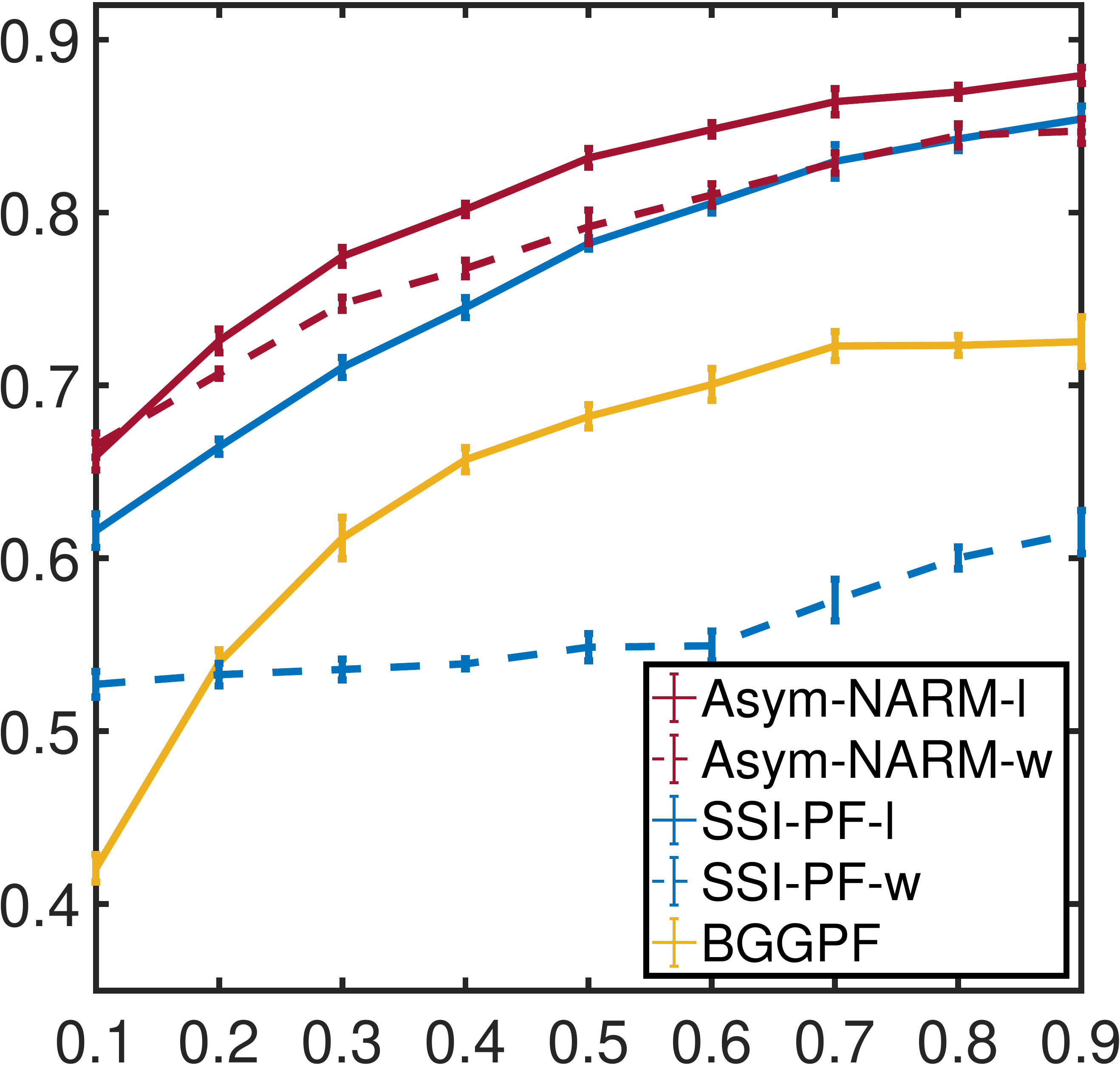}
                 
         \end{subfigure}
         \begin{subfigure}[b]{0.23\linewidth}
                 \centering
                 \caption{Aminer}
                 \label{aminer}
                 \includegraphics[width=0.98\textwidth]{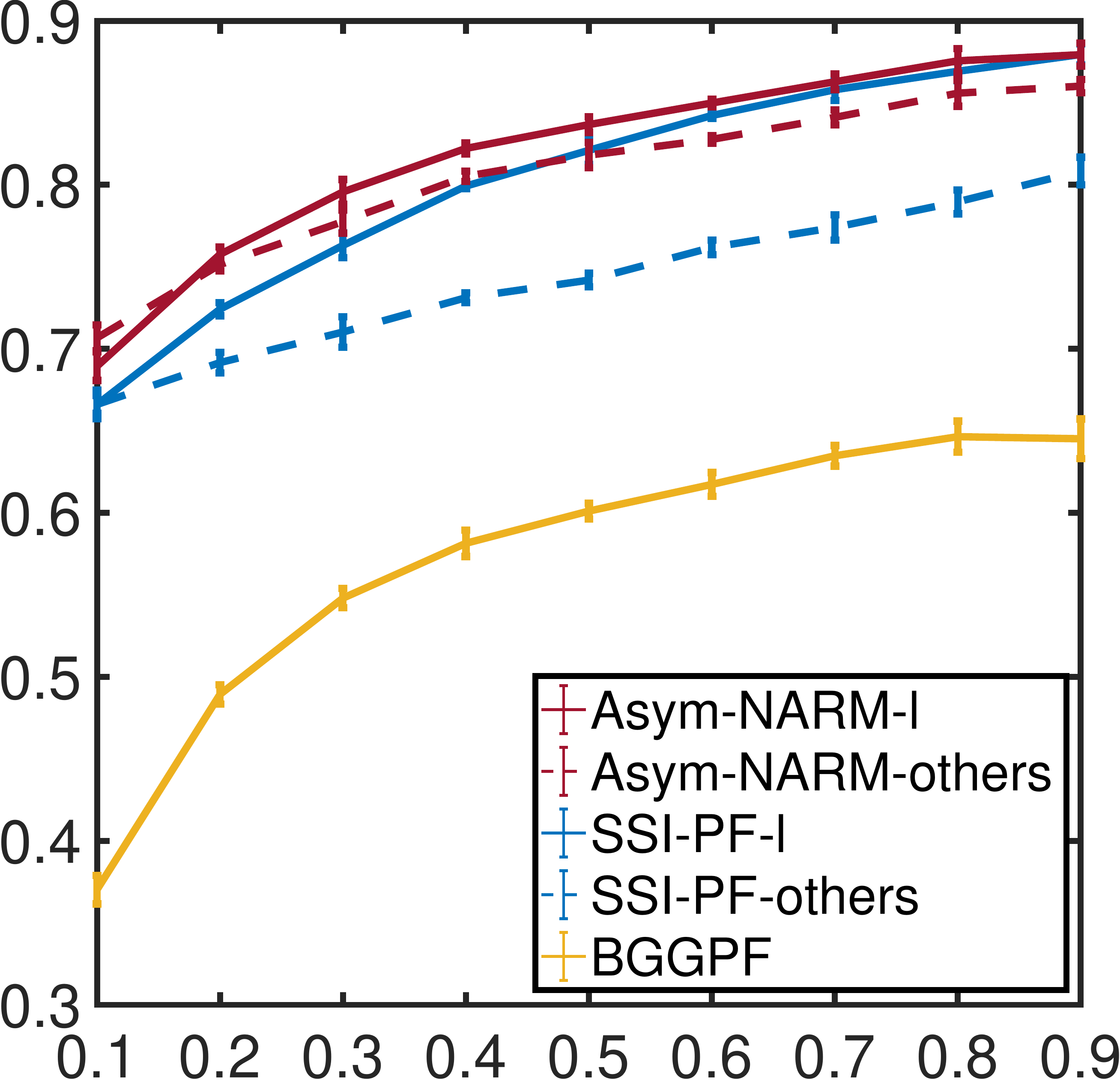}
           
         \end{subfigure}
        \begin{subfigure}[b]{0.23\linewidth}
                 \centering
                 \includegraphics[width=0.98\textwidth]{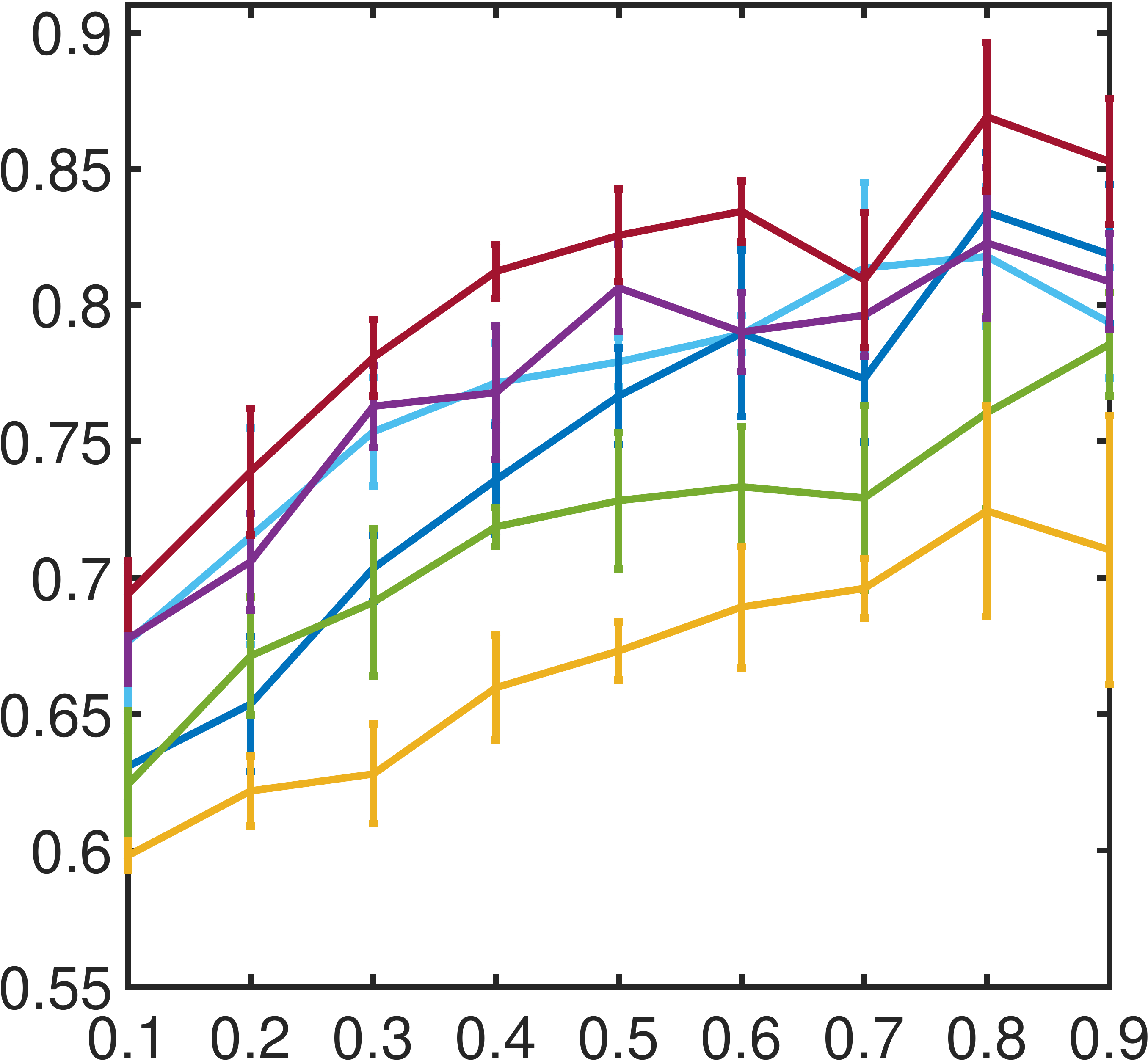}
         \end{subfigure}
         \begin{subfigure}[b]{0.23\linewidth}
                 \centering
                 \includegraphics[width=0.98\textwidth]{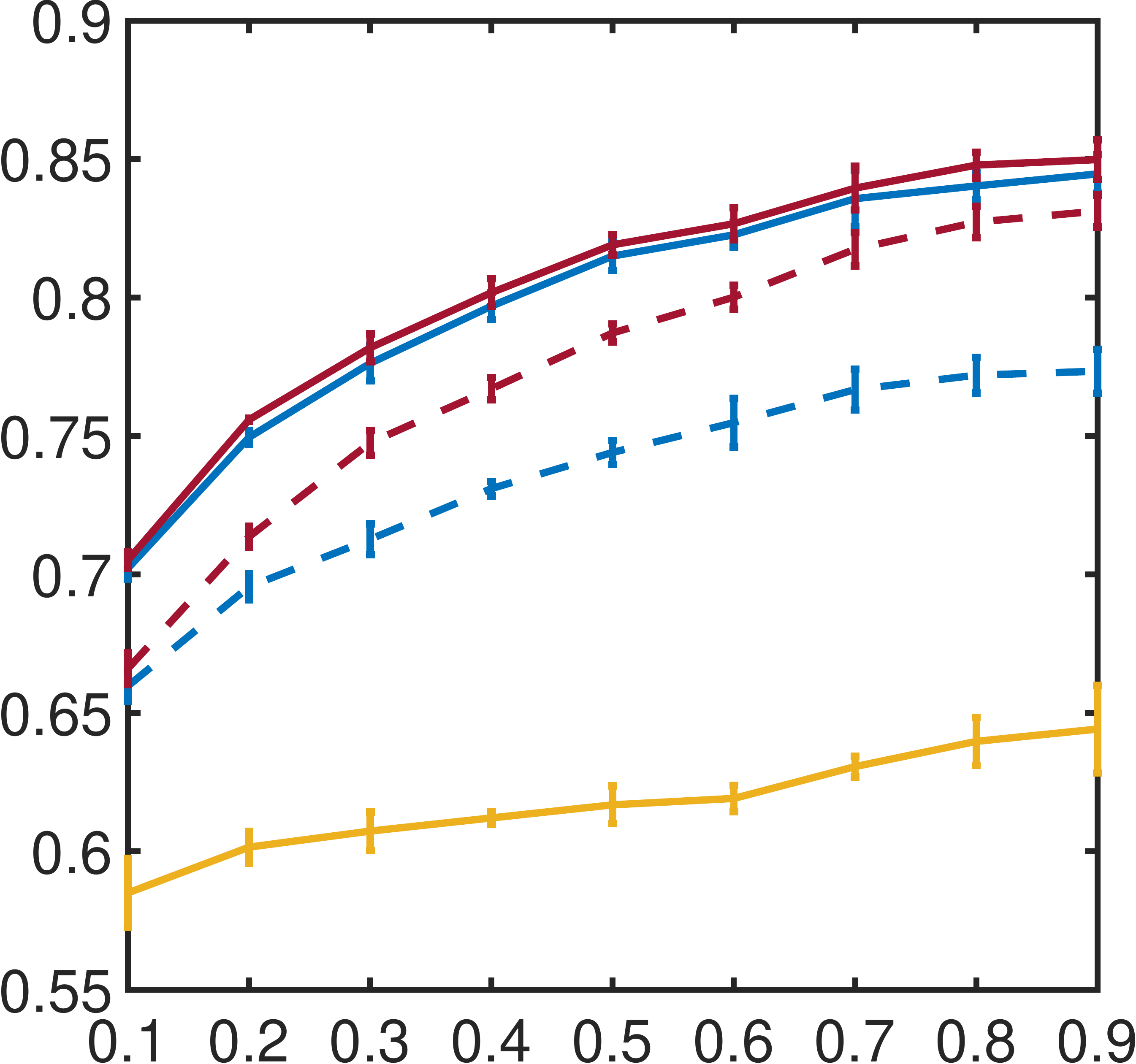}
         \end{subfigure}%
         \begin{subfigure}[b]{0.23\linewidth}
                 \centering
                 \includegraphics[width=0.98\textwidth]{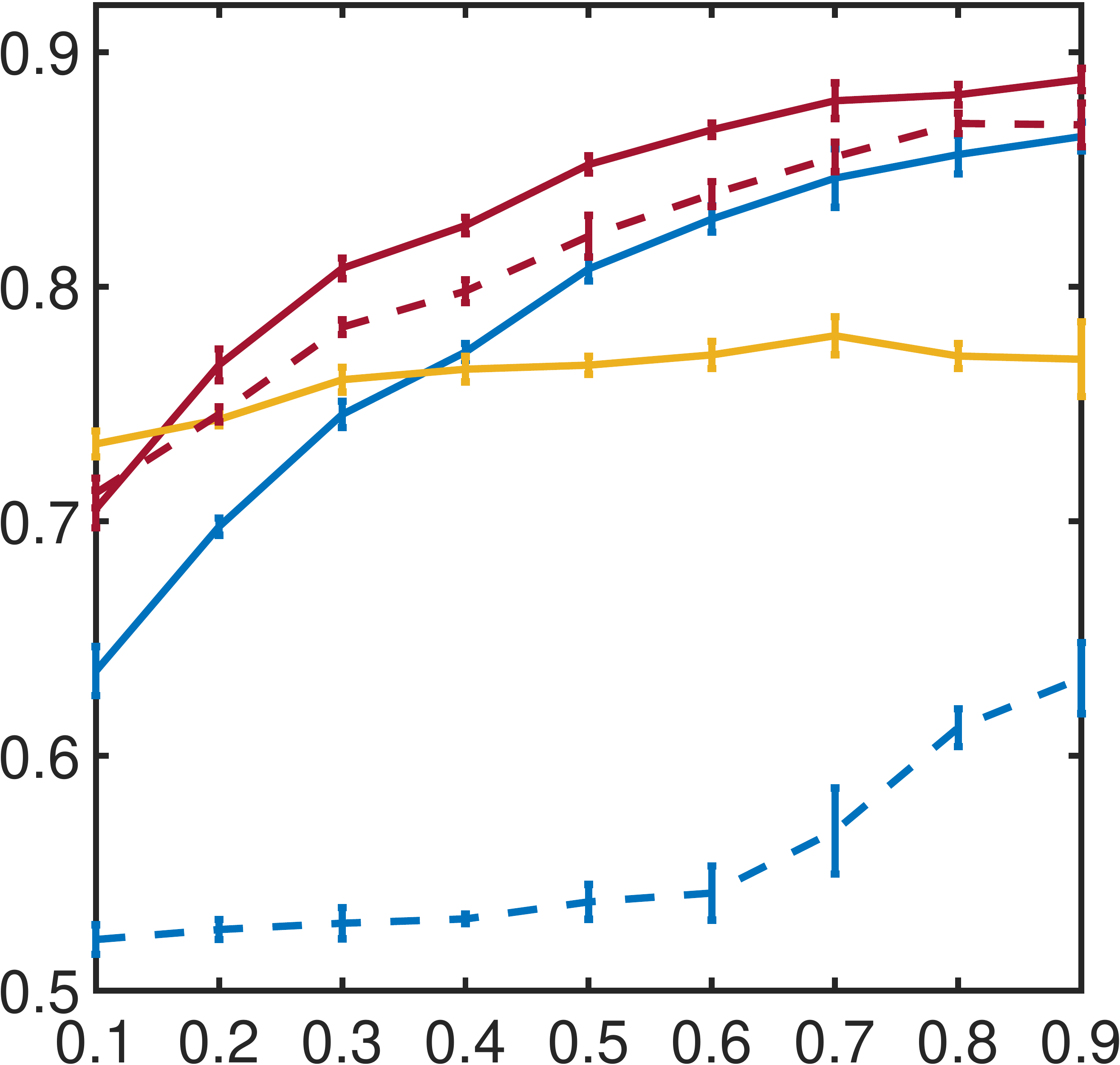}
         \end{subfigure}
         \begin{subfigure}[b]{0.23\linewidth}
                 \centering
                 \includegraphics[width=0.98\textwidth]{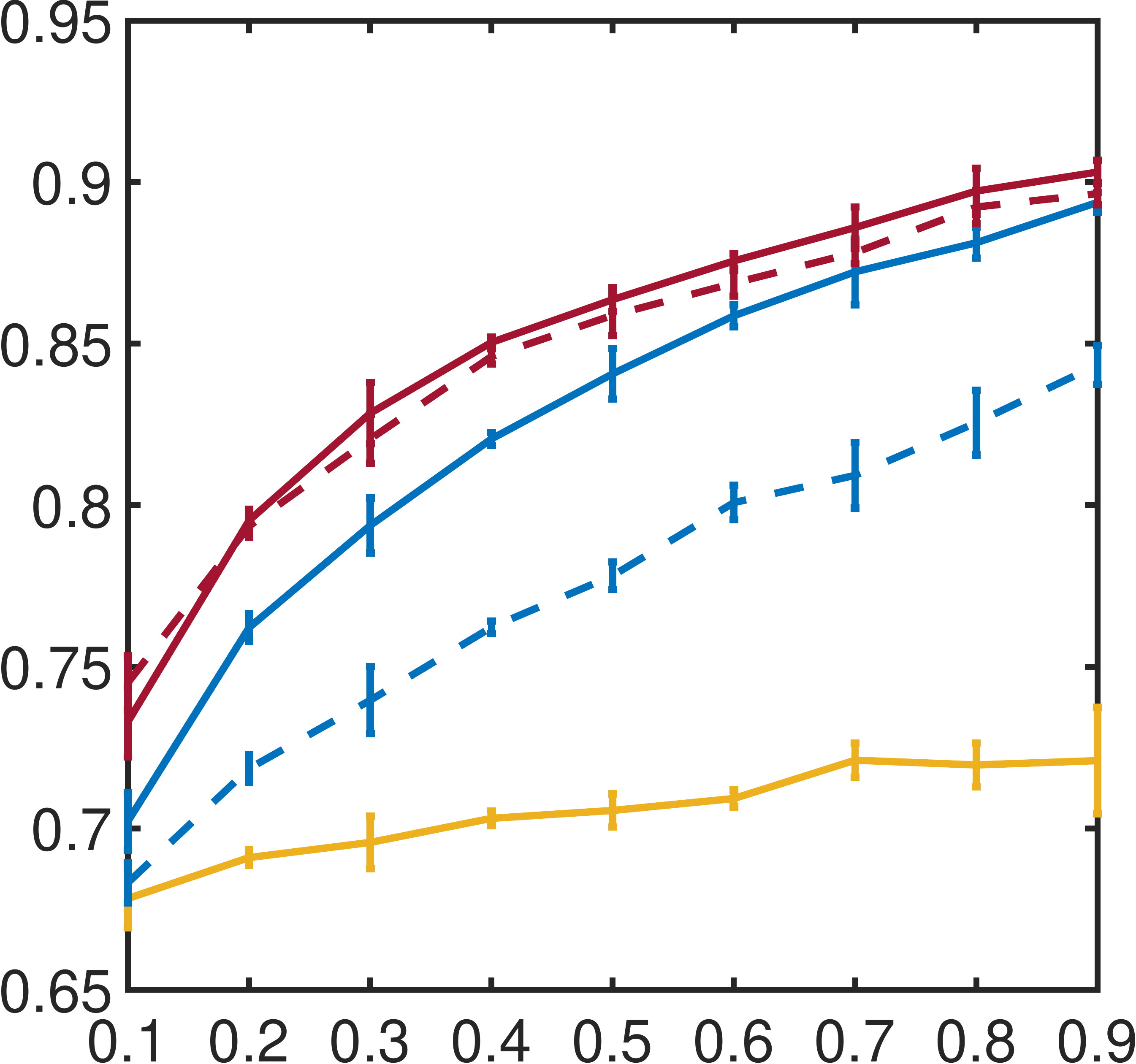}
         \end{subfigure}
\caption{The AUC-ROC (the first row) and AUC-PR (the second row) scores on the directed networks. The models with ``-l'' and ``-w'' use the labels and the words as attributes respectively. The models with ``-others'' in Aminer use the extra attributes. DMR-MMSB achieves its best performance at $K = 10$ on Lazega-advice.}
\label{fg_auc_pr_directed}
\vspace{-5mm}
\end{figure*}

\subsubsection{Results} 
The AUC-ROC/PR scores are reported in Figure
\ref{fg_auc_pr_undirected}.
Overall, our Sym-NARM model performs significantly
better than niMM, iMMM, and DMR-MMSB on all the datasets, and EPM on 3 datasets (except Facebook-ego with large training proportions).
It is interesting that the performance of EPM on Facebook-ego
gradually approaches ours when more than 30\% training data were
used. Note that Facebook-ego is
much denser than the others, which means the network information itself could be rich
enough for EPM to reconstruct the network and the node
attributes contribute less. However in general, when relational data are
highly incomplete (with less training data), our model is able to achieve improved link prediction performance.
%
 
To illustrate how side information helps, we
qualitatively compared our model with EPM and niMM by estimating the link
probabilities on NIPS234, shown in Figure~\ref{fg_reconstruct}.
With 20\% training data, EPM does not give a meaningful reconstruction
of the original network, but it starts to with more data presented.
The similarity of the authors' topics in Figure~\ref{atc} matches the original network, demonstrating the usefulness of the topics, but with some error.
Using the topics as the authors' attributes, our Sym-NARM achieves
reasonably good reconstruction of the network with only 20\% training data, 
further improving with 80\% training data. Although niMM uses the same node attributes, its
performance is not as good and is even outperformed by EPM with 80\%
training data.

\subsection{Link Prediction on Directed Networks}
\label{ss_directed}
Here we compared our \textbf{Asym-NARM} (implemented in MATLAB on top of the BGGPF code) with two
models that do not consider node attributes,
\textbf{BGGPF}~\cite{zhou2012beta} and
\textbf{iMMM},
and three node-attribute models,
\textbf{niMM},
\textbf{SSI-PF}~\cite{hu2016non} and
\textbf{DMR-MMSB}.
We used the following four datasets:
\begin{itemize}[noitemsep,leftmargin=4mm,topsep=0pt]
\item \textbf{Lazega-advice:} This dataset is a directed network with 892 links of the advice relation among the attorneys.
 The node attributes are the same as in Lazega-cowork.
\item \textbf{Citeseer:} This dataset\footnote{\label{cs_source}\url{h
ttp://linqs.umiacs.umd.edu/projects//projects/lbc/index.html}}
contains a citation network with 4591
links of 3312 papers, labelled with one of 6 categories. 
For each paper, we used both the category label and the
presence/absence of 500 most frequent words as two separate
attribute sets. We got a $3312 \times 500$ word attribute matrix
with 65674 non-zero entries. 
\item \textbf{Cora:} This dataset\footref{cs_source} contains a citation network with 5429 links
of 2708 papers in machine learning, labelled with one of 7 categories.
Similar to Citeseer, we used both the category label and the 500 most frequent words
as two separate attribute sets. We got a $2708 \times 500$ word attribute matrix
with 39268 non-zero entries. 
\item \textbf{Aminer:} The Aminer dataset \cite{tang2009social}
  contains a citation network with 2555 papers  labelled with 10 categories
  and 5967 links. We further collected information of
  each paper via the Aminer's API, including the authors' names (2597
  unique authors), abstract, venue, year, and number of citations. For the
  abstract, we extract the 5 most frequent topics for each paper in a similar way to
  NIPS234. In total, we prepared two sets of attributes:
  the labels and the others formed with the combination of all 
  collected information.
\end{itemize}

\begin{figure*}[t!p]
        \centering
         \begin{subfigure}[b]{0.23\linewidth}
                 \centering
                 \caption{Cora-hier AUC-ROC}
                 \includegraphics[width=0.98\textwidth]{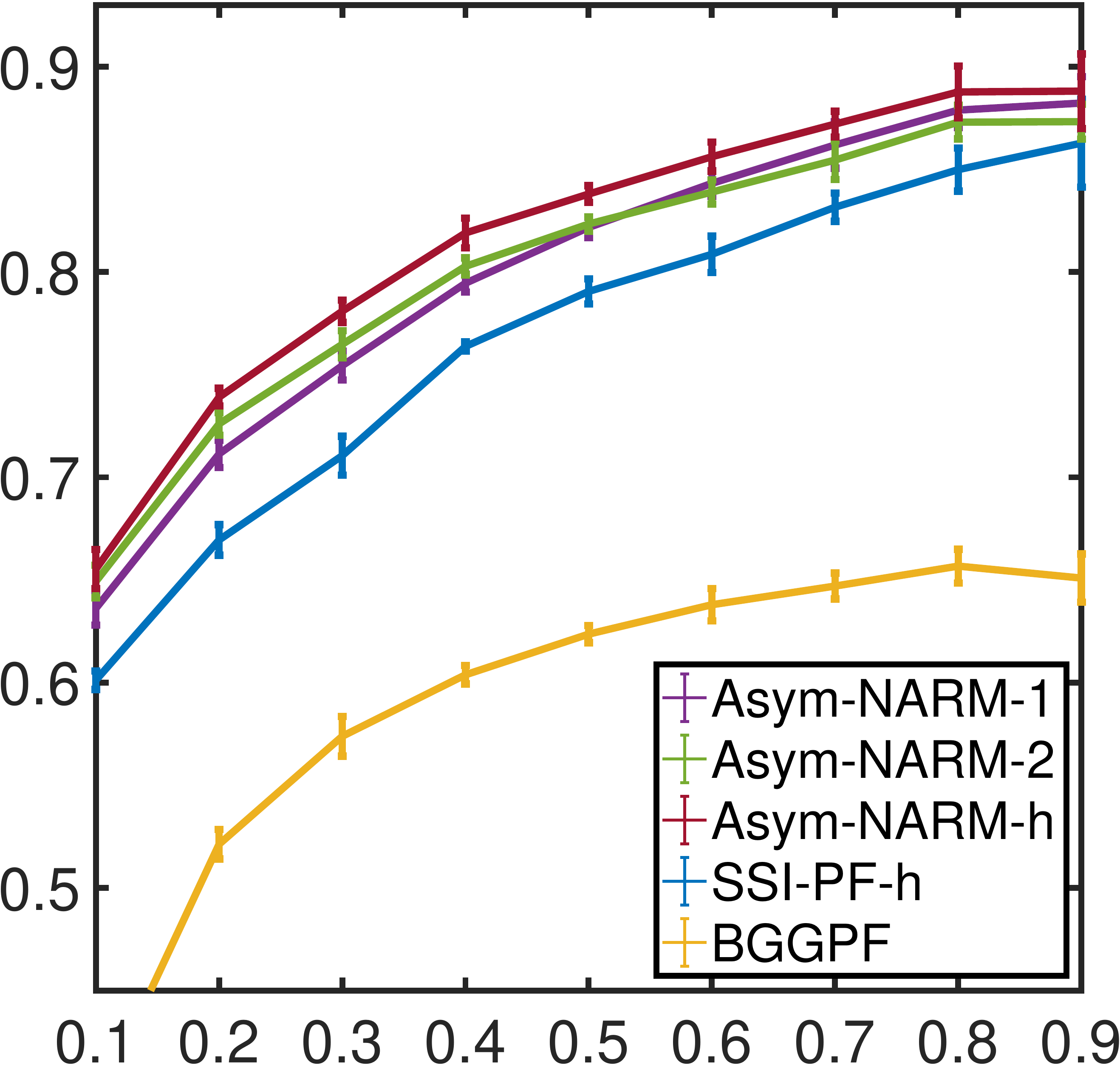}
         \end{subfigure}
         \begin{subfigure}[b]{0.23\linewidth}
                 \centering
                 \caption{Cora-hier AUC-PR}
                 \includegraphics[width=0.98\textwidth]{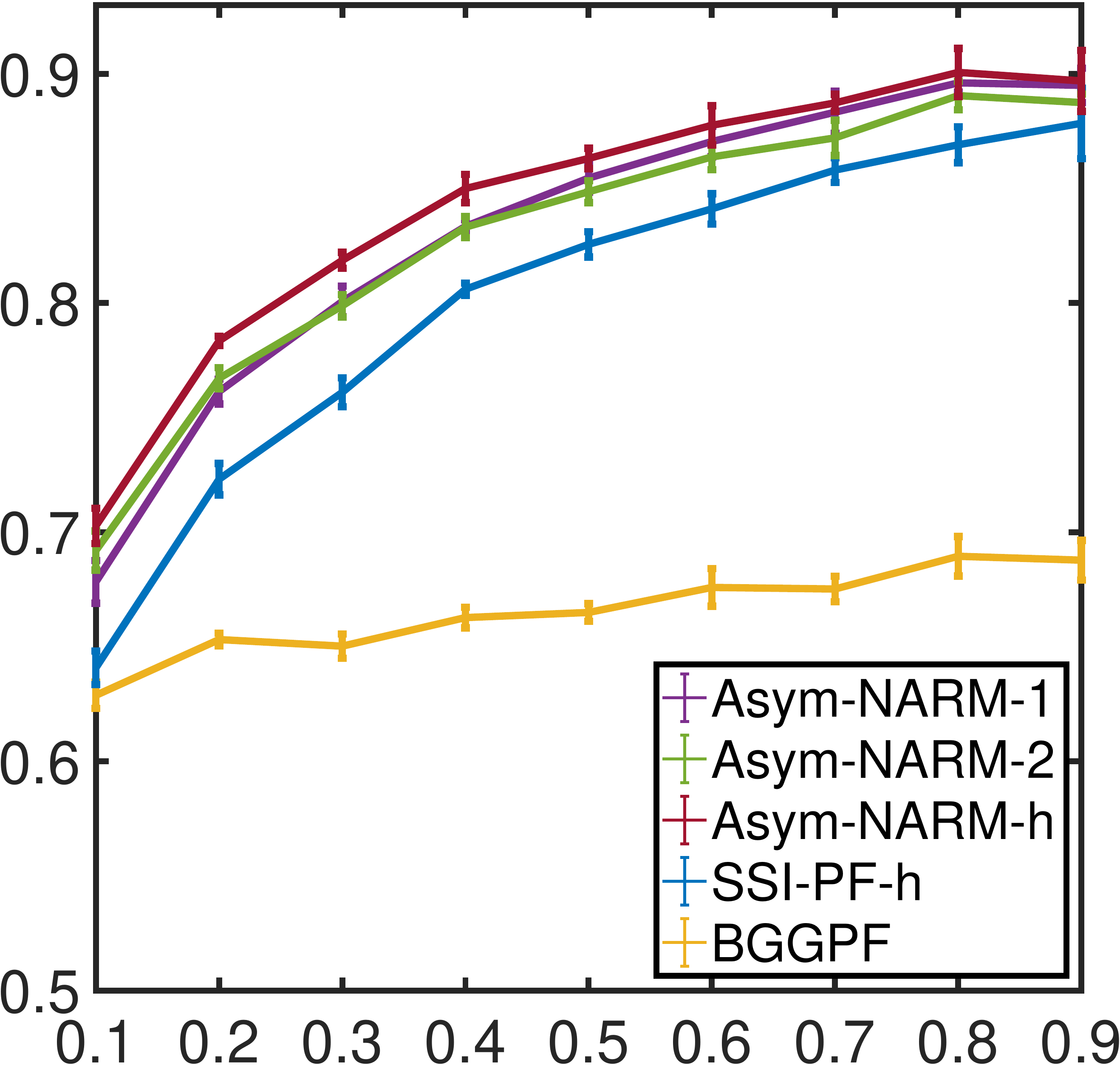}
         \end{subfigure}
         \begin{subfigure}[b]{0.23\linewidth}
                 \centering
                 \caption{Patent-hier AUC-ROC}
                 \includegraphics[width=0.98\textwidth]{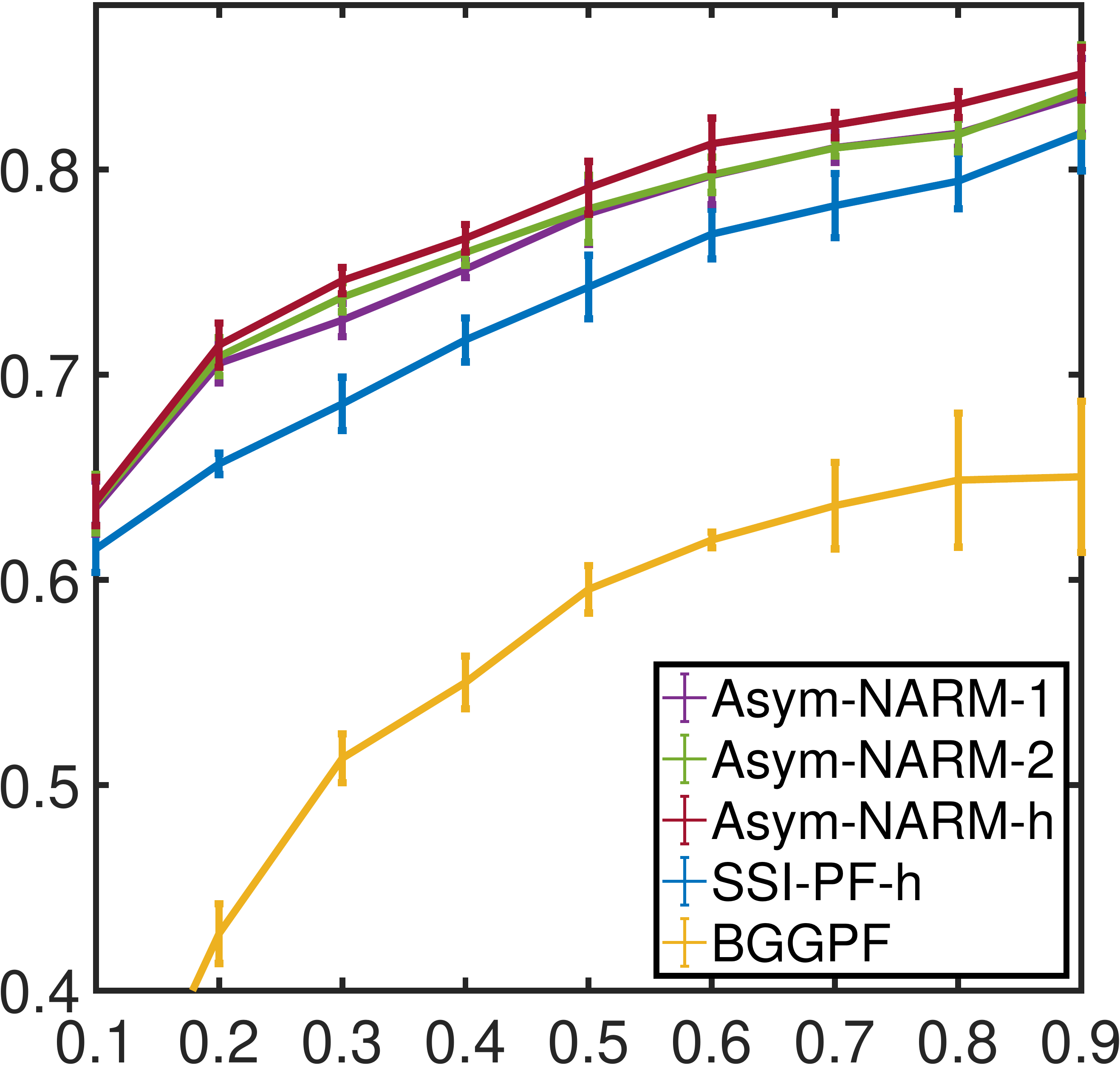}
         \end{subfigure}
        \begin{subfigure}[b]{0.23\linewidth}
                 \centering
                 \caption{Patent-hier AUC-PR}
                 \includegraphics[width=0.98\textwidth]{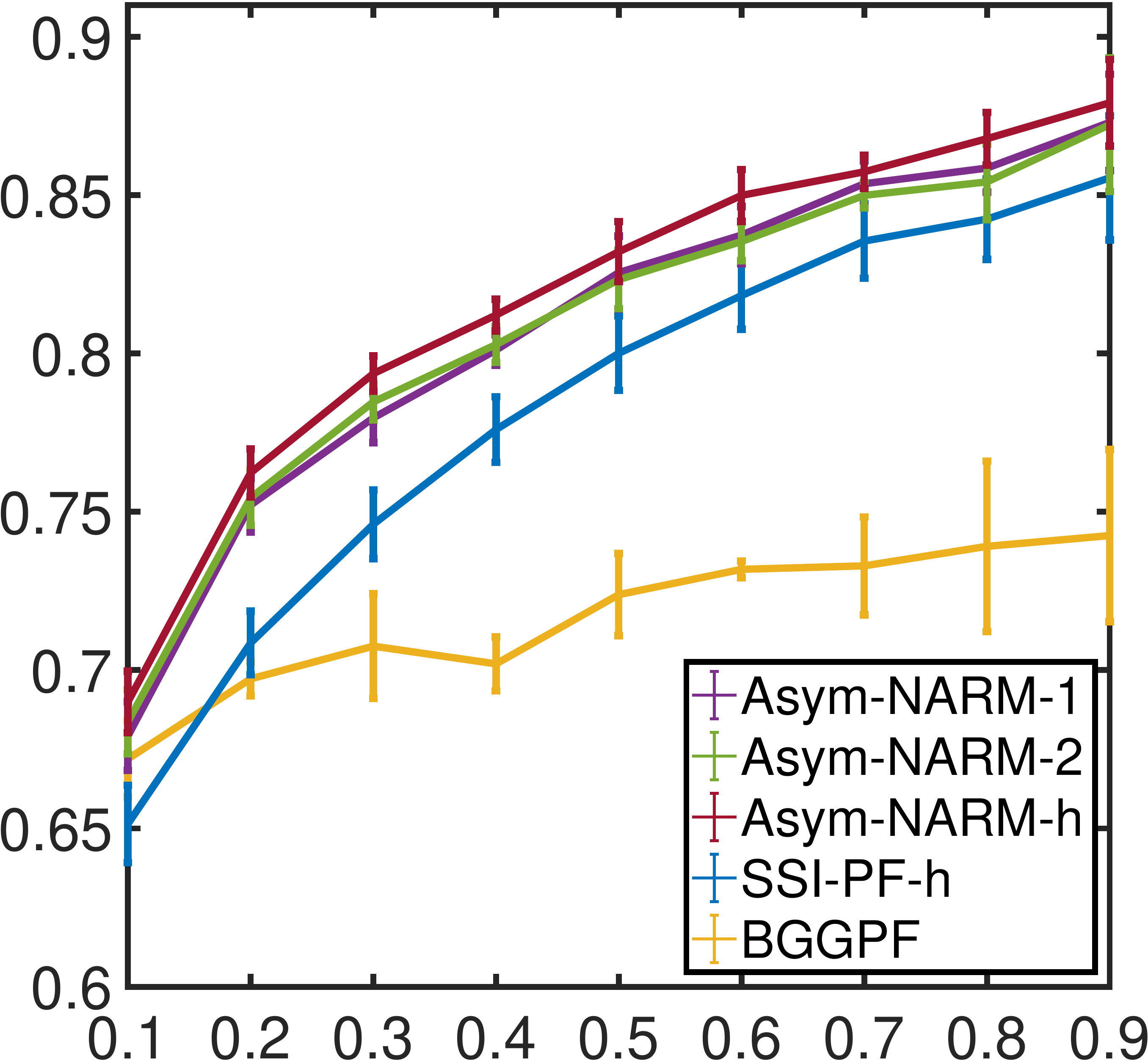}
         \end{subfigure}
\caption{The AUC-ROC and AUC-PR scores on the networks with hierarchical attributes. The models with the first level attributes only, the second level attributes only, and the hierarchical attributes are marked with ``-1'',  ``-2'', and ``-h'' respectively. 
}
\label{fg_auc_pr_hier}
\vspace{-5mm}
\end{figure*}

\subsubsection{Experimental Settings}
For fair comparison, we generated training/testing data with the
code in the SSI-PF package, which splits a network in terms of its links.  
We used the default hyper-parameter settings of BGGPF, SSI-PF, and
niMM, provided by the original authors. $K_{max}$ was set to 50 on Lazega-advice 
and 200 (same as \cite{hu2016non}) on all the other three datasets. 
For our Asym-NARM, we set $\mu_0 = 1$ and the other hyper-parameters
the same as those used in
\cite{zhou2012beta,hu2016non}. 
Following the suggestion of \citet{hu2016non}, we used
1500 MCMC iterations in total and the last 500 samples to compute the
AUC-ROC/PR scores. 
Since Citeseer, Cora, and Aminer are already too large for 
niMM, iMMM, and DMR-MMSB to produce results in reasonable time given our
computational resources, we reported their performance only on 
Lazega-advice.

\subsubsection{Results} 

Shown in Figure~\ref{lazega_advice}, Asym-NARM gains better results in terms of AUC-ROC/PR on Lazega-advice in most of the training proportions.
Overall, the node-attribute models perform better than 
the models that do not consider node attributes, showing the usefulness of node attributes. On
the other three datasets, we used different sets of attributes to
study how different attributes influence the performance of Asym-NARM
and SSI-PF.

In general, Asym-NARM performs better than SSI-PF
regardless of which set of attributes is used. The performance of SSI-PF
approaches ours in Citeseer with the labels as attributes (indicated by ``-l'').
But the gap between SSI-PF and our model becomes larger when the words are
used as attributes (indicated by ``-w'').
In Cora, SSI-PF with the words does not perform
as well as its non-node-attribute counterpart, BGGPF, indicating
it may not be as robust as our model with large sets of attributes.
To investigate this, we varied the number of the most frequent words from 10 to 500 for 
Asym-NARM and SSI-PF on Citeseer and Cora.
With more words, the AUC-ROC/PR score of SSI-PF degrades increasingly.
We further checked the prior of the node factor loadings in SSI-PF (the variable that incorporates node attributes and corresponds to $g_{i,k}$ in our model) 
and found that the coefficient of variation of each node's prior drops dramatically, 
indicating with more words, SSI-PF is failing to use the supervised information in the words.

\subsection{Link Prediction with Hierarchical Node Attributes}


Here we used two datasets with hierarchical node attributes: (1)~\textbf{Cora-hier}: a citation network with 1712 papers and 6308 links extracted from the original Cora dataset\footnote{\url{https://people.cs.umass.edu/~mccallum/data.html}}. 
The papers are labelled with one of 63 sub-areas (first level) and each sub-area belongs to one of 10 primary areas (second level), such as  ``machine learning in artificial intelligence'' and ``memory management in operating systems''; (2)~\textbf{Patent-hier}: a citation network with 1461 patents and 2141 links from the National Bureau of Economic Research where the hierarchical International Patent Classification (IPC) code of a patent is used as attributes.

The AUC-ROC/PR scores in Figure~\ref{fg_auc_pr_hier} show that our Asym-NARM with hierarchical attributes outperforms the others, which demonstrates leveraging hierarchical side information is beneficial to link prediction. Although SSI-PF also models the hierarchical attributes, its performance in these two datasets is not comparable with our model's.

\subsection{Running Time} 
\label{sec-runtime}

\setlength{\textfloatsep}{0.5cm}
\begin{table}[!th]
\centering
\vspace{-3mm}
\caption{The running time (seconds per iteration) of the compared models on Aminer. AT: the topics extracted from the abstracts. 
      All: the combination of all the attributes we have. 
      }
      \vspace{2mm}
\label{tb_rt_feature}
\scalebox{0.8}{
\begin{tabular}{|c|c|c|c|c|c|}
\hline
Attr   &  \specialcell{Non-zeros\\ \& attr size}   &  \specialcell{\textbf{Asym-}\\\textbf{NARM}} & SSI-PF & niMM & \specialcell{DMR-\\MMSB\\$K=50$} \\ \hline
Label     &  \specialcell{2660\\2555*10} &    \textbf{0.26}      & 0.48   & 134.11 & 89.12  \\ \hline
AT    &  \specialcell{12775\\2555*100}  &    \textbf{0.29}      & 0.87   & 135.22   & 126.44 \\ \hline
Authors &  \specialcell{5647\\2555*2597}  &    \textbf{0.33}      & 2.99   & 136.41  & - \\ \hline
All    &  \specialcell{31273\\2555*3058}  &    \textbf{0.51}      & 5.21   & 136.14   & - \\ \hline
\end{tabular}%
}
\vspace{-3mm}
\end{table}

In this section, we compare the running time of the models for directed
networks (all implemented in MATLAB and running on a desktop with
3.40 GHz CPU and 16GB RAM). Using 80\% data for training, the running
time for Asym-NARM, SSI-PF, and niMM on
Aminer with different sets of node attributes is reported in
Table~\ref{tb_rt_feature}. Note DMR-MMSB did not complete with ``Authors'' and ``All'' due to our computational resources. Asym-NARM is about 10
times faster than SSI-PF with all the attributes and about 2 times faster with the labels.
Thus Asym-NARM is more efficient, especially with large sets of attributes,
supporting the complexity analysis in Table~\ref{tb_cc}. 

\section{Conclusion}
\label{sec-dis}
As a summary of the experiments, Asym/Sym-NARM achieved better link prediction performance with faster inference. While EPM, a non-node-attribute model, performed well on nearly complete networks, it degraded with less training data. niMM and DMR-MMSB,
extensions to MMSB with the logistic-normal transform, had similar results to Sym-NARM
but scaled inefficiently. SSI-PF's performance and scalability were not as good as Asym-NARM in the presented cases with flat and hierarchical attributes and it was less effective with larger numbers of attributes.

Thus NARM is a comparatively simple yet effective and efficient way of
incorporating node attributes, including hierarchical attributes,
for relational models with Poisson likelihood.
This leads to improved link prediction and matrix
completion for less complete relational data of
both directed and undirected networks. With the efficient
inference, our models can be used to model large sparse relational
networks with node attributes. 

NARM can easily be extended
to multi-relational networks such as \cite{hu2016topic} and topic
models with document and word attributes, which is left for our future work.


\bibliography{icml}
\bibliographystyle{icml2017}

\end{document}